\documentclass[10pt,twocolumn,letterpaper]{article}

\usepackage[pagenumbers]{cvpr} % To force page numbers, e.g. for an arXiv version

\usepackage{graphicx}
\usepackage{amsmath}
\usepackage{amssymb}
\usepackage{booktabs}

\usepackage[accsupp]{axessibility}  % Improves PDF readability for those with disabilities.

\usepackage[pagebackref,breaklinks,colorlinks]{hyperref}

\usepackage[capitalize]{cleveref}
\crefname{section}{Sec.}{Secs.}
\Crefname{section}{Section}{Sections}
\Crefname{table}{Table}{Tables}
\crefname{table}{Tab.}{Tabs.}

\begin{document}

\title{GIRAFFE HD: A High-Resolution 3D-aware Generative Model}

\vspace{-0.2in}
\author{
Yang Xue\textsuperscript{1} \quad 
Yuheng Li\textsuperscript{2} \quad 
Krishna Kumar Singh\textsuperscript{3} \quad 
Yong Jae Lee\textsuperscript{2}  
\vspace{0.05in}
\\
\textsuperscript{1}UC Davis
\quad \textsuperscript{2}UW--Madison
\quad \textsuperscript{3}Adobe Research
}

\twocolumn[{%
	\maketitle
	\vspace{-0.8cm}
	\renewcommand\twocolumn[1][]{#1}
	\begin{center}
		\centering
		\includegraphics[width=0.95\textwidth]{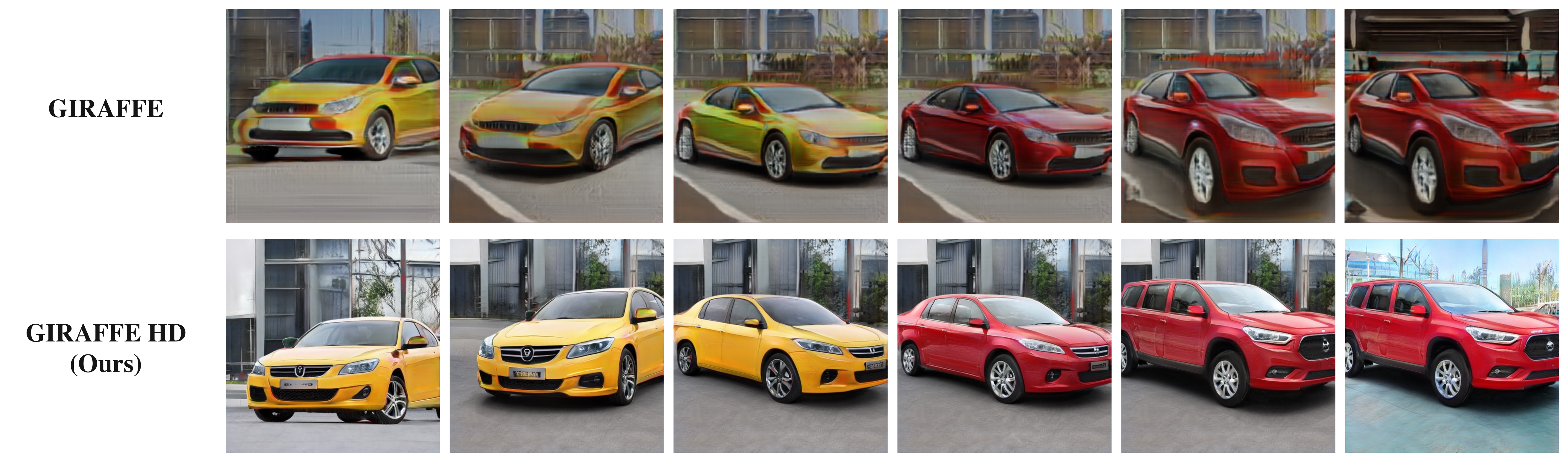}
		\captionof{figure}{Our model, GIRAFFE HD, inherits all of GIRAFFE's~\cite{Niemeyer2020GIRAFFE} 3D controllability---change in camera viewpoint, object translation, scale, rotation, appearance, shape, and background---while generating higher quality, higher resolution images. Moreover, it achieves better foreground-background disentanglement; e.g., when changing the car's shape (fourth and fifth columns), notice how parts of the road and building in the background change in the GIRAFFE images, whereas they remain constant in ours.}
		\label{fig:teaser}
	\end{center}
}]

%%%%%%%%% ABSTRACT
\begin{abstract}
\vspace{-5pt}
3D-aware generative models have shown that the introduction of 3D information can lead to more controllable image generation.  In particular, the current state-of-the-art model GIRAFFE~\cite{Niemeyer2020GIRAFFE} can control each object's rotation, translation, scale, and scene camera pose without corresponding supervision.  However, GIRAFFE only operates well when the image resolution is low. We propose GIRAFFE HD, a high-resolution 3D-aware generative model that inherits all of GIRAFFE's controllable features while generating high-quality, high-resolution images ($512^2$ resolution and above).  The key idea is to leverage a style-based neural renderer, and to independently generate the foreground and background to force their disentanglement while imposing consistency constraints to stitch them together to composite a coherent final image.  We demonstrate state-of-the-art 3D controllable high-resolution image generation on multiple natural image datasets.
\end{abstract}

%%%%%%%%% BODY TEXT
\vspace{-5pt}
\section{Introduction}
% \label{sec:intro}

In image generation, two of the most important objectives are image realism and controllability. Style-based GANs (i.e., StyleGAN~\cite{Karras-CVPR2018} and its variants~\cite{Karras2019stylegan2,Karras-NeurIPS2021}) can generate high-resolution, photorealistic images. However, while their latent style code design provides a level of disentanglement and controllability in 2D space (e.g., color and shape changes), their lack of explicit 3D information makes it difficult to impose \emph{3D-level control} over the generated image content. Meanwhile, the recent NeRF~\cite{mildenhall2020nerf} based GANs~\cite{Schwarz-neurips2020,Niemeyer2020GIRAFFE,chanmonteiro2020piGAN} have shown that explicit modeling of the scene in 3D space conditioned on camera pose can enable effective 3D-level control. However, the computationally expensive nature of 3D representations has limited current 3D-aware generative models from directly learning and rendering images in high resolutions.

GIRAFFE~\cite{Niemeyer2020GIRAFFE} is the current state-of-the-art 3D-aware generative model for both image realism and controllability. It models the foreground and background as two separate 3D objects, uses volume rendering to render the combined 3D features into low-resolution 2D feature maps, and finally uses a neural renderer to further render the feature maps into higher resolution images.  These design choices enable GIRAFFE to change the background's appearance independent of the foreground, translate or rotate the foreground object in 3D, and change the foreground object's shape and color. However, the neural renderer is specifically designed to provide only spatially small refinements in order to avoid entangling global scene properties and losing controllability.  Thus, it is significantly less powerful than style-based renderers, and hence the highest image resolution that GIRAFFE can generate is $256^2$.

In this work, we propose a two-stage style-based 3D-aware generative model that inherits all of GIRAFFE's controllability while generating high-quality, high-resolution images (up to $1024^2$ resolution); see Fig.~\ref{fig:teaser}. Our design is motivated by three key observations when replacing GIRAFFE's neural renderer with a style-based neural render (based on StyleGAN2~\cite{Karras2019stylegan2}): 1) Using the style renderer to upsample the volume-rendered low-res 2D feature maps leads to high-quality, high-resolution image generations while still preserving controllability over the foreground object's 3D properties (translation, rotation). However, due to its high capacity, the style renderer 2) now gains full control over color as well as some control over shape, and 3) entangles and loses controllability over the foreground and background features (i.e., changing the foreground color/shape also changes the background color/shape).  

In order to regain controllability over the foreground and background, we generate them independently using two different style-based renderers and combine them into a coherent image by imposing geometric and photometric compatibility constraints that eliminate inconceivable combinations.
Furthermore, to disentangle color and shape, we exploit the well-known emergent properties of StyleGAN, namely that the early layers control coarse shape, mid layers control fine-grained shape, and later layers control color.  Specifically, we inject the shape code into the 3D feature generator as well as the early layers of the style renderer to control shape, and the appearance code into the later layers of the style renderer to control color. 

\vspace{-10pt}
\paragraph{Contributions.}
Our approach, GIRAFFE HD, preserves the 3D controllability of GIRAFFE, including independent control over foreground and background, while generating much higher-resolution and higher-quality images (up to $1024^2$ vs.~GIRAFFE's $256^2$).  We validate our approach on multiple natural image datasets (CompCar~\cite{compcars}, FFHQ~\cite{karras-cvpr2019}, AFHQ Cat~\cite{Starganv2}, CelebA-HQ~\cite{Karras-iclr2018}, LSUN Church~\cite{Yu2015LSUNCO}) and demonstrate better foreground-background disentanglement and image realism compared to GIRAFFE in higher resolution domains.  Finally, we perform ablation studies to justify the different design choices for our model.

%-------------------------------------------------------------------------

\section{Related Work}
\label{sec:related}

\paragraph{3D-aware image synthesis.} In recent years, using implicit neural representations to represent 3D scenes and volume render into 2D images has shown great potential~\cite{chen-cvpr2019,mescheder-cvpr2019,Saito-iccv2019pifu,genova-iccv2019,Chibane-NeurIPS2020,Chiyu-CVPR2020,Sitzmann-NeurIPS2020,chen-cvpr2021}. For example, NeRF~\cite{mildenhall2020nerf} can effectively learn the 3D geometry using multiple images of a scene from different viewpoints and generate new images from new viewpoints. NeRF-based 3D-aware GANs~\cite{Schwarz-neurips2020,Niemeyer2020GIRAFFE,chanmonteiro2020piGAN, chan2021pi} condition the neural representations on sample noise or appearance/shape codes to represent different 3D scenes with a single network. This improvement also enables these models to be trained on unstructured image collections, as opposed to images from a single scene.  Our GIRAFFE HD builds upon this line of work; in particular, it extends GIRAFFE~\cite{Niemeyer2020GIRAFFE} to higher-resolution image domains while retaining all its 3D understanding capabilities. 

\vspace{-10pt}
\paragraph{High-resolution image synthesis.} Generative Adversarial Networks (GANs)~\cite{goodfellow-nips2014,radford-iclr16,brock2018large} can generate photorealistic images, and the state-of-the-art for high-resolution image synthesis are style-based GANs~\cite{Karras-CVPR2018,Karras2019stylegan2,Karras-NeurIPS2021}. By injecting style codes~\cite{Huang-ICCV2017} to the network, these models achieve not only high-resolution outputs but also some level of feature disentanglement (e.g., pose, shape, lighting). Some recent work~\cite{Li2021CollagingCG,lewis2021tryongan} have demonstrated that using StyleGAN2 as a neural renderer can effectively upsample the low-resolution feature maps produced by another network into high-resolution images. We leverage the StyleGAN2 architecture as our model's neural renderer to generate high-resolution 2D images ($512^2$ resolution and higher).

\vspace{-10pt}
\paragraph{Disentanglement and controllability.}
Generative models that learn disentangled representations~\cite{hinton-icann2011,bengio-tpami2013,chen-nips16,higgins-iclr2017,denton-nips17,hu-cvpr18,Karras-CVPR2018,WonkwangECCV2020,PeeblesECCV2020,partgan} provide extra control in their generations, for example, the ability to control different factors in the scene (e.g., object pose, shape, appearance). However, most methods only operate in the 2D domain without considering the 3D structure of the objects/scenes.

\begin{figure*}[t!]
    \centering
    \includegraphics[width=0.98\textwidth]{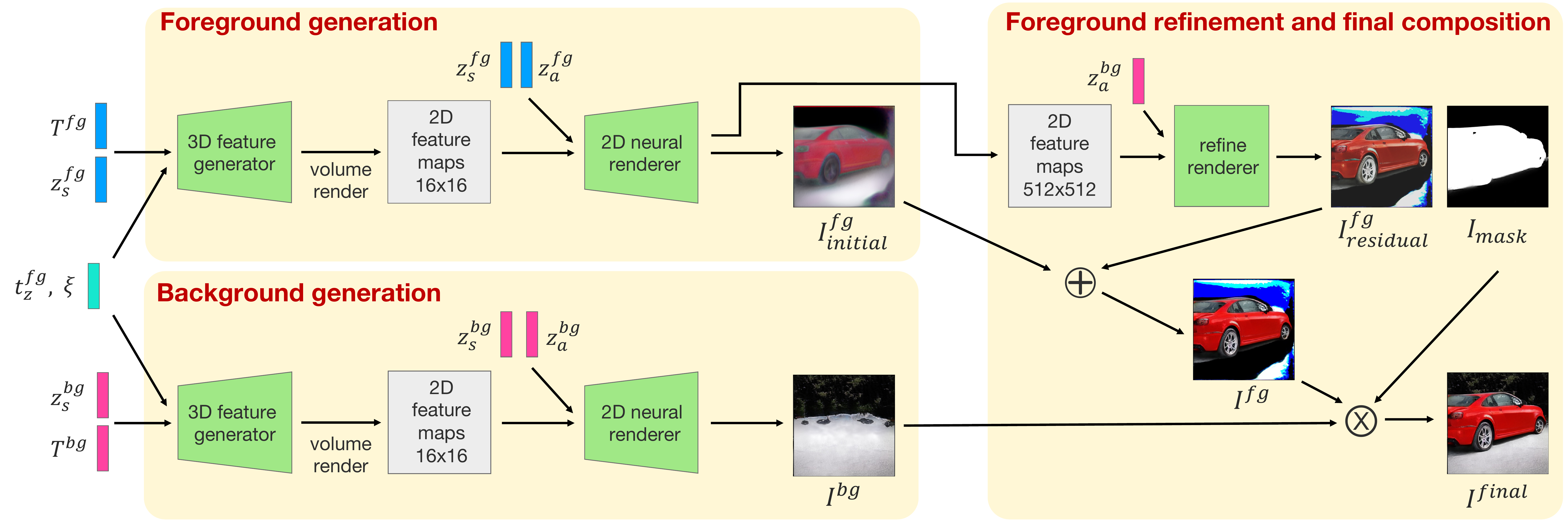}
    \caption{\textbf{GIRAFFE HD Architecture.} Our model independently generates the foreground and background and uses a generated mask to composite the final image. The camera pose $\xi$ and foreground object's z-translation $t_z^{fg}$ are shared between the foreground and background 3D feature generators to ensure geometric compatibility. To ensure photometric compatibility, the refinement renderer injects environment information conditioned on background appearance $z_a^{bg}$ to generate the foreground residual image $I^{fg}_{residual}$, which is added to the initial output image of the foreground 2D neural renderer $I^{fg}_{initial}$ to form the final foreground image $I^{fg}$.  During evaluation, latent codes are strategically injected into the 2D neural renderers to ensure disentanglement over appearance and fine-grained shape.}
    \label{fig:approch}
\end{figure*}

Among generative models that learn 3D disentanglement~\cite{henzler2019escaping,BlockGAN2020,nguyen2019hologan,Niemeyer2020GIRAFFE,schwarz2020graf}, GIRAFFE~\cite{Niemeyer2020GIRAFFE} is the current state-of-the-art.  It represents a 3D scene as a composition of foreground and background 3D objects, which enables it to disentangle object shape, appearance, position, camera viewpoint, as well as the foreground and background during image synthesis. However, we observe that this disentanglement comes as a trade-off to image quality -- replacing GIRAFFE's low capacity neural renderer with a style-based renderer leads to high-resolution synthesis but at the loss of foreground-background disentanglement. Several supervised methods exploit StyleGAN's style-based disentanglement properties to control the generation process~\cite{abdal-tog2021,wu2021stylespace,patashnik2021styleclip}. However, they have yet to demonstrate accurate foreground-background disentanglement, which suggests that the vanilla StyleGAN architecture has a limitation when it comes to foreground-background disentanglement. The most intuitive and reliable foreground-background disentanglement methods are two-stage image generators, which generate the foreground and background independently and use 2D composition to form the final output image~\cite{yang-iclr17,Singh2019FineGAN,bielski-nips2019,Li2020MixNMatch}. However, these methods fall short in terms of image quality, presumably due to a lack of explicit foreground and background information sharing and compatibility constraints. Our approach also generates the foreground and background in two separate stages, but it imposes explicit geometric and photometric compatibility constraints.  This leads to accurate foreground-background disentanglement in the high-resolution image domain.

%-------------------------------------------------------------------------

\vspace{-1pt}
\section{Approach}
\label{sec:method}
\vspace{-1pt}

Given an image collection containing a single object category (e.g., cars), our goal is to learn a 3D-aware image generation model that generates photo-realistic, high-resolution images while also providing 3D-level control without human supervision.  To this end, our architecture builds upon GIRAFFE~\cite{Niemeyer2020GIRAFFE}, but replaces its low-capacity neural render with a StyleGAN2~\cite{Karras2019stylegan2} based renderer, and has two separate parallel streams to generate separate foreground and background images to enforce their disentanglement.  We create the final output by combining the foreground and background while imposing compatibility constraints to ensure a coherent image; see Fig.~\ref{fig:approch}.

\subsection{Background on GIRAFFE}

\paragraph{Generative neural feature fields.} 
GIRAFFE~\cite{Niemeyer2020GIRAFFE} represents a 3D scene with a neural feature field~\cite{mildenhall2020nerf,Niemeyer2020GIRAFFE}, which is a continuous function $F$ that maps a 3D location $x \in \mathbb{R}^3$ and a 2D camera viewing direction $d \in \mathbb{S}^2$ to a density $\sigma \in \mathbb{R}^+$ and an appearance feature $f \in \mathbb{R}^{M_f}$.  It uses an MLP to learn $F$, and conditions it on $z \sim \mathcal{N} (0, I)$ so that each $z$ corresponds to a different 3D scene:
\begin{equation}
\vspace{-1pt}
  F_\theta: (\gamma^{L_x}(x), \gamma^{L_d}(d), z) \mapsto (\sigma, f)
%   \label{eq:pos-encoding}
\vspace{-1pt}
\end{equation}
where $\theta$ indicates the network parameters, $\gamma$ is a positional encoding~\cite{Tancik-NeurIPS2020} which maps the 5D input $(x, d)$ into a higher dimensional space, and $L_x$ and $L_d$ are the positional encoding dimensions of $x$ and $d$, respectively. 

\paragraph{3D object representation.}
GIRAFFE represents the foreground and background objects using two separate MLPs associated with separate affine transformations $T^{fg}$ and $T^{bg}$, respectively, sampled from a dataset-dependent distribution $T = \{s, t, R\}$, where $s, t \in \mathbb{R}^3$ are scale and translation parameters, and $R \in SO(3)$ is a rotation matrix. The affine transformation $T$ transforms the scene's world location to the object's local location for each object:
\begin{equation}
  \kappa(x) = R \cdot sE \cdot x + t
%   \label{eq:pos-encoding}
\end{equation}
where $E$ is the $3\times 3$ identity matrix.  This representation enables 3D object-level control.

\vspace{-10pt}
\paragraph{Volume rendering.}
For a given camera pose $\xi$, let $\mathit{\{x_{j}\}}^{N_s}_{j=1}$ be $N_s$ sample points along camera ray $d$ for a given pixel. Then
\vspace{-1pt}
\begin{equation}
  (\sigma_{j}, f_{j}) = F_\theta(\gamma^{L_x}(\kappa^{-1}(x_{j})), \gamma^{L_d}(d), z).
%   \label{eq:pos-encoding}
\vspace{-1pt}
\end{equation}

Let $\delta_{j} = \mathit{\Vert x_{j+1} - x_{j} \Vert} ^2$ denote the distance between neighboring sampled points, $\alpha_j = 1 - e^{-\sigma_j \delta_j}$ denote the alpha value for $x_j$, and $\tau_{j} = {\prod}_{i=1}^{j-1} 1 - \alpha_j$ denote the transmittance along the ray.  Pixel feature vector $f$ can then be computed using numerical integration:
\begin{equation}
  f = {\sum}^{N_s}_{j=1} \tau_{j} \alpha_{j} f_{j}
%   \label{eq:pos-encoding}
\end{equation}

For efficiency, the rendered feature images are at $16^2$ resolution.  The volume-rendered feature map $f_{vol}$ can then be processed by a neural renderer (i.e., a convnet) to output the final RGB image.

Note that in GIRAFFE, the foreground and background's 3D object representations are composed into a single 3D scene representation and volume-rendered into a single 2D feature representation. However, in our approach, we will independently volume render the foreground and background's 3D representations, as explained in detail next.

\subsection{Neural style rendering}

In GIRAFFE, the neural renderer is purposely designed to be simple and provide only spatially small refinements to the volume-rendered feature maps, in order to avoid entangling global scene properties and losing controllability. With its default renderer, the highest resolution it can generate is $256^2$. 

In order to generate higher-resolution ($\ge$ $512^2$) outputs, we first replace GIRAFFE's default neural renderer with one based on StyleGAN2~\cite{Karras2019stylegan2}.  Specifically, we take all the blocks of StyleGAN2 starting from $16^2$ resolution to convert the volume-rendered $16^2$ resolution 2D feature maps $f_{vol}$ into a higher resolution image $I$. As in StyleGAN2, we also use a mapping network to map $z \sim \mathcal{N} (0, I)$ to latent code $w$:
\vspace{-5pt}
\begin{align}
  \psi_\theta:& z \mapsto w \\
  \pi^{render}_\theta:& (f_{vol}, w) \mapsto I
%   \label{eq:pos-encoding}
\end{align} 

While our style renderer leads to higher-resolution outputs, we observe several behavioral distinctions compared to GIRAFFE's default renderer.  First, the model now loses its ability to independently control the foreground and background.  Second, the 3D representation no longer fully controls the object's shape. Though it still determines overall coarse shape, the earlier stages of the style renderer gain finer level control over the shapes since the 3D representation is volume-rendered to 2D feature maps at a much lower resolution than the final image. Third, the 3D representation almost does not control color at all. Instead, the control of color is transferred to the later stages of the style renderer. These behaviors resemble those of vanilla StyleGAN2. 

To regain independent control over foreground and background, and to better disentangle object color and shape, we make the following design choices. First, instead of compositing the scene at the 3D level and then rendering it into a single final 2D image, we first render the foreground and background independently into two 2D images and then perform 2D composition to get the final image.  Second, unlike GIRAFFE, which conditions the 3D representation on the object's shape code $z_{s}$ as well as its appearance code $z_{a}$, we remove the dependency of each point's feature $f$ on $z_{a}$. Instead, during training, we perform style mixing in the style renderer (as described in StyleGAN2) with $w_{s} = \psi_\theta(z_{s})$ and $w_{a} = \psi_\theta(z_{a})$. During evaluation, we inject $w_{s}$ into the earlier stages and $w_{a}$ into the later stages of the style renderer (we vary the injection index depending on final image resolution). This way of injecting codes enables our model to disentangle color and shape finely.

\subsection{Enforcing foreground-background consistency}

In order to combine the separately generated foreground and background images into a coherent final image, we need to impose geometric and photometric consistency between the foreground and background objects.  Geometric consistency requires the foreground and background objects to obey physical world rules; for example, objects in the same image have to share the same viewing perspective, or a car cannot be floating in the air. Photometric consistency requires the foreground and background objects to appear to reside in the same environment by sharing the same lighting, hue or saturation, etc.  To this end, we devise two mechanisms to satisfy the two consistency requirements: position sharing and environment sharing.

\vspace{-10pt}
\paragraph{Position sharing.} 
We activate position sharing when the background contains a ground surface that the foreground rests on (e.g., car on road or church on land). We simplify the problem and assume that the ground can only be a plane surface. With this simplification, simply placing the bottom of the foreground object onto the ground and then synchronizing the viewing angles for the foreground and background object will satisfy the geometric consistency requirement.  We perform this by copying the foreground object's $z$-translation and view perspective to the background object's $z$-translation and view perspective. In this way, the generated background can actively accommodate all foreground objects.

For datasets where the foreground object is not expected to rest on a ground surface (e.g., frontal human faces), we only synchronize the viewing perspective without sharing the $z$-translation between foreground and background.

\vspace{-10pt}
\paragraph{Environment sharing.}

Besides enforcing geometric consistency between the foreground and background objects, we also need to ensure photometric consistency; i.e., the foreground object should naturally be immersed into the environment created by the background. To this end, we designate the background appearance latent code $w_a^{bg}$ to encode the scene environment configuration. Our refinement network consists of several layers of style-based convolutions. It takes as input the foreground feature maps $f_{out}^{fg}$, which are also used to render the initial foreground image, and $w_a^{bg}$ as the style code, and outputs the foreground image residual $I_{residual}^{fg}$ and foreground object mask $I_{mask}$:
\begin{equation}
  \pi^{refine}_\theta: (f_{out}^{fg}, w_a^{bg}) \mapsto (I_{residual}^{fg}, I_{mask})
%   \label{eq:pos-encoding}
\end{equation}

We add the foreground image residual to the initial foreground image to get the final foreground image:
\begin{equation}
  I^{fg} = I_{initial}^{fg} + I_{residual}^{fg}
%   \label{eq:pos-encoding}
\end{equation}

\begin{figure}[t!]
    \centering
    \includegraphics[width=0.46\textwidth]{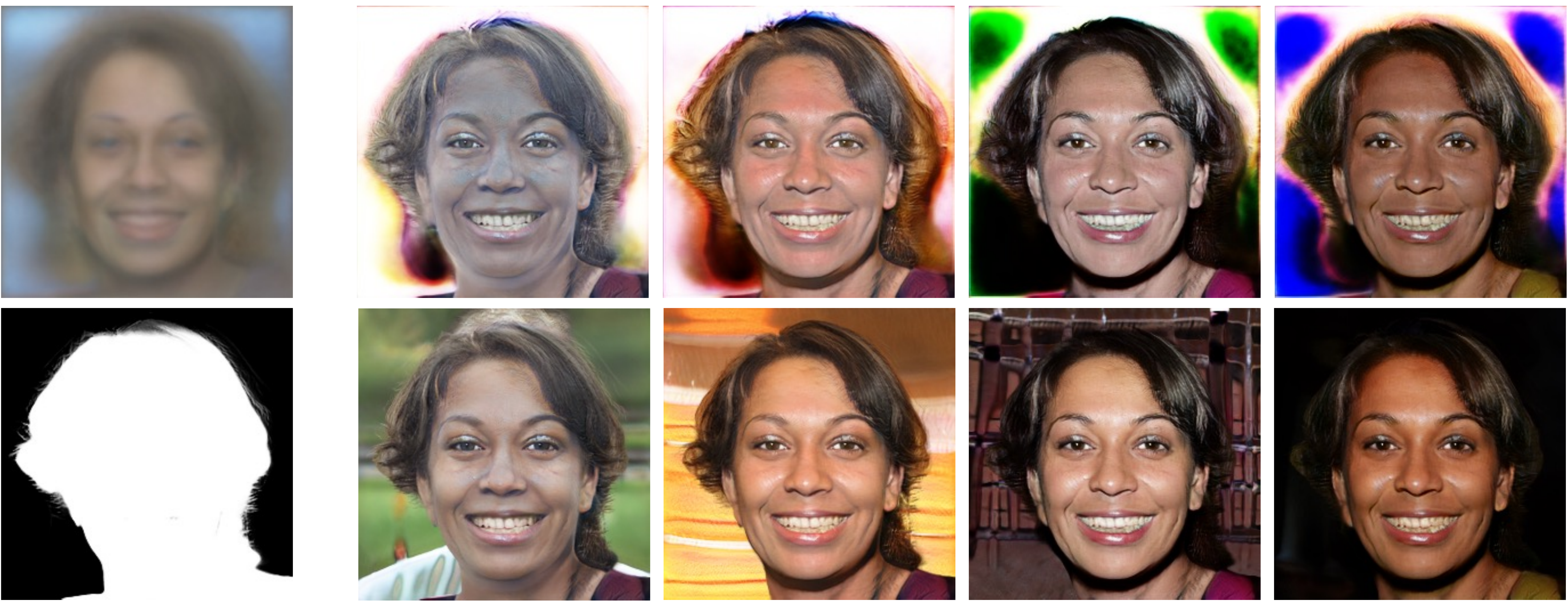}
    %\vspace{-5pt}
    \caption{\textbf{Enforcing Photometric Consistency.} First column: initial foreground image $I^{fg}_{initial}$ and mask $I_{mask}$. First row: foreground residuals $I^{fg}_{residual}$; second row: final images. Notice how $I^{fg}_{residual}$ changes based on the background so that the refined foreground $I^{fg}$ becomes more compatible with it.}
    \label{fig:ffhq_bg}
    %\vspace{-0.1in}
\end{figure}

We observe that the initial foreground image already determines the foreground object's true appearance. The refinement operation only adjusts the shading/shine of the foreground without altering its true appearance; see Fig.~\ref{fig:ffhq_bg}.

\subsection{Compositing the final image}

Finally, we use the foreground object mask $I_{mask}$ generated by the refinement network to perform alpha composition of the foreground and background images:
\begin{equation}
  I^{final} = (1-I_{mask}) \cdot I^{bg} + I_{mask} \cdot I^{fg}
  \label{eq:final}
\end{equation}
where $I^{final}$ is our model's final generated image.

Like GIRAFFE, we can generalize our model to generate multiple foreground objects. To do this, we first render the background and the foregrounds as described previously. We then compute occlusion relations between the foreground objects (by ordering based on their depth i.e., $x$-translation). Finally, we recursively perform 2D composition (Eqn.~\ref{eq:final}) from the furthest to the nearest foreground object, where in each recursive iteration, the 2D composition result becomes the new background image.

\subsection{Training}

\paragraph{Discriminator.}
We use the same residual discriminator as StyleGAN2~\cite{Karras2019stylegan2}.

\vspace{-10pt}
\paragraph{Training.}
During training, we follow~\cite{Niemeyer2020GIRAFFE} and sample latent codes $z_a^{k}, z_s^{k} \sim \mathcal{N}$, $T^k \sim p_T$ and $\xi \sim p_\xi$, where $k \in \{\text{fg},\text{bg}\}$, $p_\xi$ and $p_T$ are uniform distributions over dataset-dependent camera elevation angles and valid object transformations, respectively.

\vspace{-10pt}
\paragraph{Objectives.}
Our overall objective function is:
\begin{equation}
  L = L_{GAN} + \frac{\lambda}{2}L_{R1} + \beta_1 L_{bbox} + \beta_2 L_{cvg} + \beta_3 L_{bin}
%   \label{eq:pos-encoding}
\end{equation}
where $\lambda = 10$, and $\beta_1, \beta_2, \beta_3$ are dataset specific.  To enforce image realism, we use the non-saturating GAN objective $L_{GAN}$~\cite{goodfellow-nips2014} with R1 regularization $L_{R1}$~\cite{Mescheder-ICML2018}.% adopted from StyleGAN2~\cite{Karras2019stylegan2}.

In addition, we employ three auxiliary losses to guide 2D foreground-background disentanglement: bounding box containment loss $L_{bbox}$, foreground coverage loss $L_{cvg}$, and mask binarization loss $L_{bin}$. $L_{cvg}$ and $L_{bin}$ are adapted from \cite{bielski-nips2019}. Since the style neural renderer is very powerful on its own, the three auxiliary losses are necessary to prevent either the foreground or background renderer from generating the entire image by itself. 

See supp.~for the full expression of the loss functions, including how the sampled appearance, shape, and camera/transformation latent codes are used. 
\vspace{-10pt}
\paragraph{Bounding box containment loss.}
Each randomly sampled foreground affine transformation $T$ determines a 3D bounding box within which the foreground object should reside. After projecting both the 3D foreground object and the 3D bounding box to 2D, the 2D foreground object should still reside in the 2D bounding box. Our bounding box containment loss minimizes the mean of the foreground object mask values that fall outside the 2D bounding box:
\begin{equation}
  L_{bbox} = \frac{1}{|S|}{\sum}_{i\in S} I_{mask}[i] \cdot (1 - I_{2Dbbox}[i])
%   \label{eq:pos-encoding}
\end{equation}
where $S$ is the set of all pixels in final image.  This loss prevents the foreground renderer from generating background features.

\vspace{-10pt}
\paragraph{Foreground coverage loss.}
This is a hinge loss on the mean mask value to ensure that the foreground is not empty:
\begin{equation}
  L_{cvg} = \max(0, \eta - \frac{1}{|S|}{\sum}_{i\in S} I_{mask}[i])
%   \label{eq:pos-encoding}
\end{equation}
where $\eta$ is the minimum coverage threshold. This prevents the background renderer from generating the entire image.

\vspace{-10pt}
\paragraph{Mask binarization loss.}
This loss encourages binarization (i.e., 0 or 1 values) of the mask:
\begin{equation}
  L_{bin} = \frac{1}{|S|}{\sum}_{i\in S} \min(I_{mask}[i]-0, 1-I_{mask}[i]).
%   \label{eq:pos-encoding}
\end{equation}

\section{Experiments}

We evaluate GIRAFFE HD's 3D controllability, with a focus on foreground and background disentanglement and their geometric/photometric consistency. We also evaluate how well it generates high-quality, high-resolution images.  Finally, we perform ablation studies to evaluate its different components and losses.

\begin{figure*}[t!]
    \centering
    \includegraphics[width=.99\textwidth]{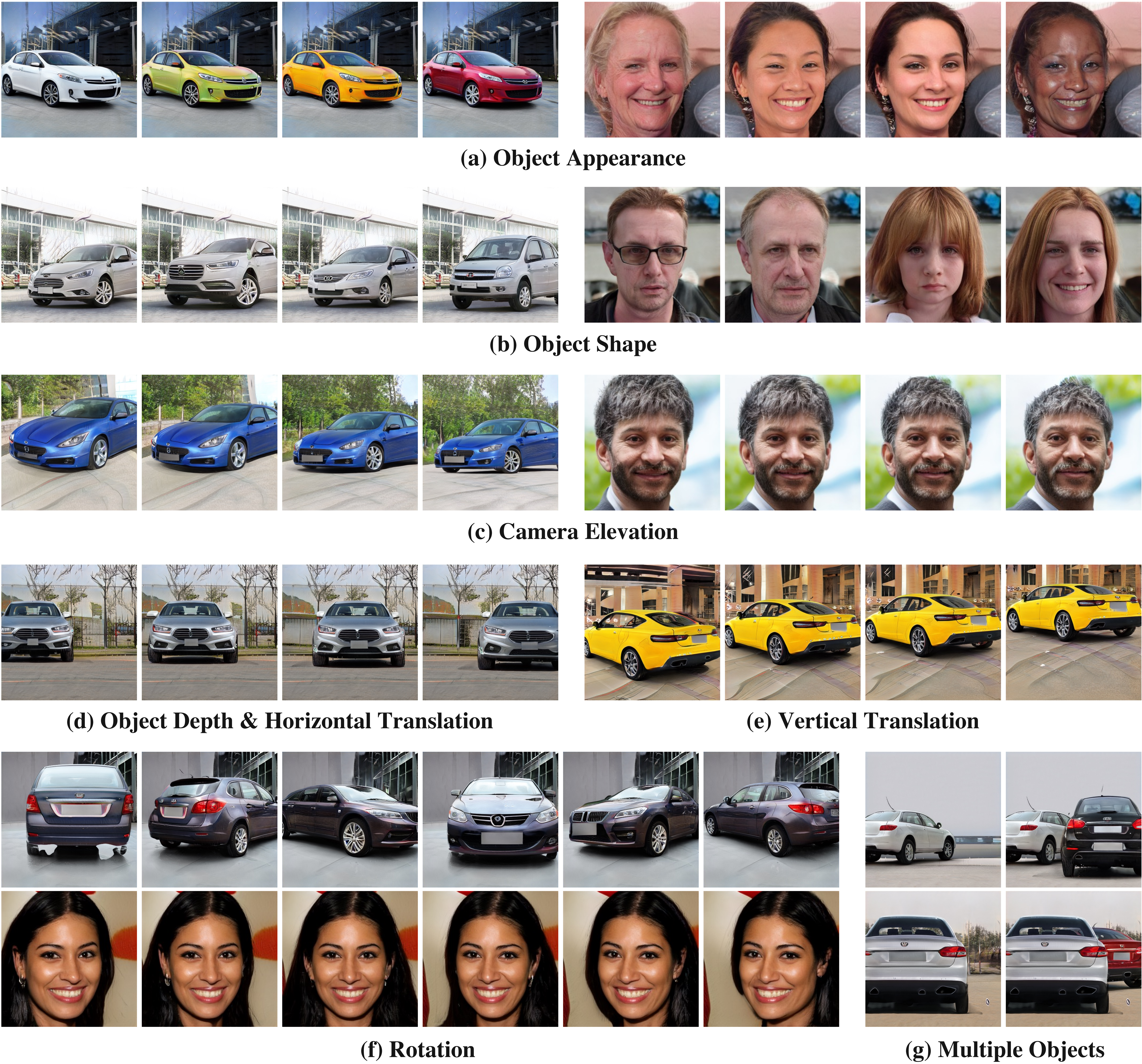}
    \caption{\textbf{3D Controllability.} GIRAFFE HD preserves all of GIRAFFE's 3D controllable features while generating images at significantly higher quality and resolution. Importantly, compared to GIRAFFE, our background remains more consistent when the foreground changes.}
    \label{fig:3dcontrol}
\end{figure*}

\vspace{-10pt}
\paragraph{Imp.~details.}
Foreground and background generative neural radiance fields are MLPs with ReLU activations. We use 8 layers with hidden dimension of 256/64 (foreground/background respectively), density of 1, and feature head of dimensionality $M_f = 256$ for the MLPs. We sample $N_s = 64$ points along each ray and render 2D feature maps at $16^2$ pixels. Both foreground and background's shape and appearance codes are 256 dimensions. We use 4 layer MLPs to map noise vectors to style renderers' latent codes. Refine renderer has 4 style-based convolution layers. We use a minimum coverage threshold of 0.2. Foreground and background images go through a tanh activation before 2D composition. Foreground mask goes through a sigmoid activation before 2D composition. We use Adam optimizer with a learning rate of 0.0005 and batch size of 16.

\vspace{-10pt}
\paragraph{Baselines.}
We compare to GIRAFFE \cite{Niemeyer2020GIRAFFE}, pi-GAN \cite{chan2021pi}, GRAF \cite{schwarz2020graf}, HoloGAN \cite{nguyen2019hologan}, and HoloGAN w/o 3D Conv, a HoloGAN variant proposed in \cite{schwarz2020graf} for higher resolutions.% 

\vspace{-10pt}
\paragraph{Datasets.}
We evaluate on five high-resolution single-object real-world datasets used in GIRAFFE \cite{Niemeyer2020GIRAFFE}: CompCar~\cite{compcars}, FFHQ~\cite{karras-cvpr2019}, AFHQ Cat~\cite{Starganv2}, CelebA-HQ \cite{Karras-iclr2018}, LSUN Church~\cite{Yu2015LSUNCO}. 

\vspace{-10pt}
\paragraph{Metrics.}
We use FID~\cite{heusel2017gans} to quantify image quality. We use 20,000 real and fake samples to calculate the FID score in order for a direct comparison to~\cite{Niemeyer2020GIRAFFE}. 

To quantify foreground-background disentanglement, we propose the mutual background similarity (MBS) metric. It measures the consistency in background between two generated images that are supposed to share the same background. A low MBS indicates more consistent background between the pair of images. For each generated image, we randomly sample an operation that should change its foreground (i.e., a combination of change in scale, x, y-translation, rotation, shape, and appearance) without altering the background, then perform that operation to generate a new image. We then use a pretrained DeepLabV3 ResNet101~\cite{chen2017rethinking} semantic segmentation model to compute the background mask for each image, and multiply the two masks to get a single mutual background mask. The image pair's MBS is computed as the fraction of pixels inside the mutual background area whose RGB value has changed. We compute the final MBS as the mean of 10,000 image pairs' MBS's $\times 10^2$. Please refer to the supp. for details.

\begin{table}[t!]
  \centering
  \resizebox{0.49\textwidth}{!}{
  \begin{tabular}{@{}lccccc@{}}
    \toprule
                & Cat & CelebA-HQ     & FFHQ         & CompCar      & Church\\
    \midrule
    HoloGAN~\cite{nguyen2019hologan}$^\dagger$    &  -  &  61 & 192 & 34 & 58\\
    w/o 3D Conv~\cite{schwarz2020graf}$^\dagger$ &  -  &  33 & 70  & 49 & 66\\
    GRAF~\cite{schwarz2020graf}$^\dagger$   & -  & 49   & 59  & 95 & 87\\  
    GIRAFFE~\cite{Niemeyer2020GIRAFFE}$^\dagger$     & 33.39 &  21 & 32  & 26 & 30\\  
    pi-GAN~\cite{chan2021pi}      & 38.92  & 36.27      & 43.19        & 64.01        & 56.80\\
    Ours        & 12.36  & 8.09       & 11.93         & 7.22         & 10.28\\
    \bottomrule
  \end{tabular}
  }
  \caption{\textbf{\boldmath{$256^2$} Resolution Image Quality.} We report the FID score (↓) for all methods. $\dagger$ scores (except Cat) taken from~\cite{Niemeyer2020GIRAFFE}.}
  \label{tab:256fid}
\end{table}

\begin{table}[t!]
  \centering
  \footnotesize
    \begin{tabular}{@{}lcc@{}}
    \toprule
                & CompCar $512^2$         & FFHQ $1024^2$\\
    \midrule
    GIRAFFE~\cite{Niemeyer2020GIRAFFE}     & 40.81         &  70.08 \\  
    Ours        & 8.36    &            10.13 \\ 
    \bottomrule
  \end{tabular}
  \caption{\textbf{\boldmath{$512^2$} and \boldmath{$1024^2$} Resolution Image Quality.} We report the FID score (↓) for GIRAFFE and GIRAFFE HD.}
  \label{tab:512fid}
\end{table}

\begin{figure}[t!]
    \centering
    \includegraphics[width=0.48\textwidth]{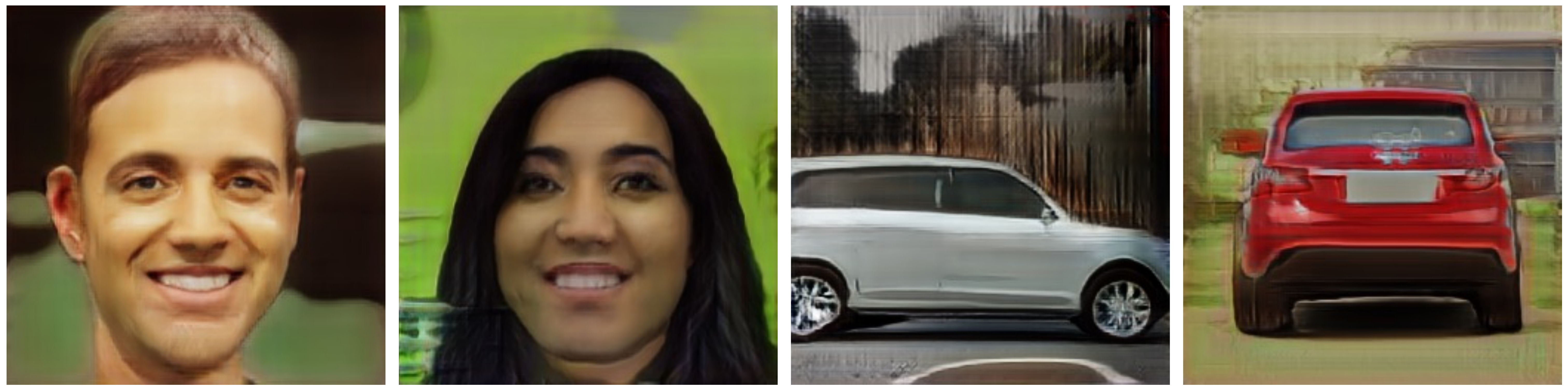}
    \caption{\textbf{GIRAFFE~\cite{Niemeyer2020GIRAFFE} image generations.} We show these to enable direct comparison. (More comparisons in the supp.)}
    \label{fig:giraffesamples}
\end{figure}

\subsection{Image generation quality}

We first evaluate the quality of GIRAFFE HD's generated images.  Since it is demonstrated in~\cite{Niemeyer2020GIRAFFE} that GIRAFFE can reliably operate at $256^2$ resolution, we start our comparison at $256^2$ against all baselines (Table~\ref{tab:256fid}). We then compare to GIRAFFE at the higher resolutions of $512^2$ for CompCar dataset and $1024^2$ for FFHQ dataset (Table~\ref{tab:512fid}).  Our method outperforms the baselines in terms of image quality by a large margin.  This can be attributed largely to our style-based neural renderer, which is able to model finer details than GIRAFFE's low-capacity neural renderer.  

\subsection{Controllable scene generation}
\begin{figure}[t!]
    \hspace{-0.5cm}
    \centering
    \includegraphics[width=0.5\textwidth]{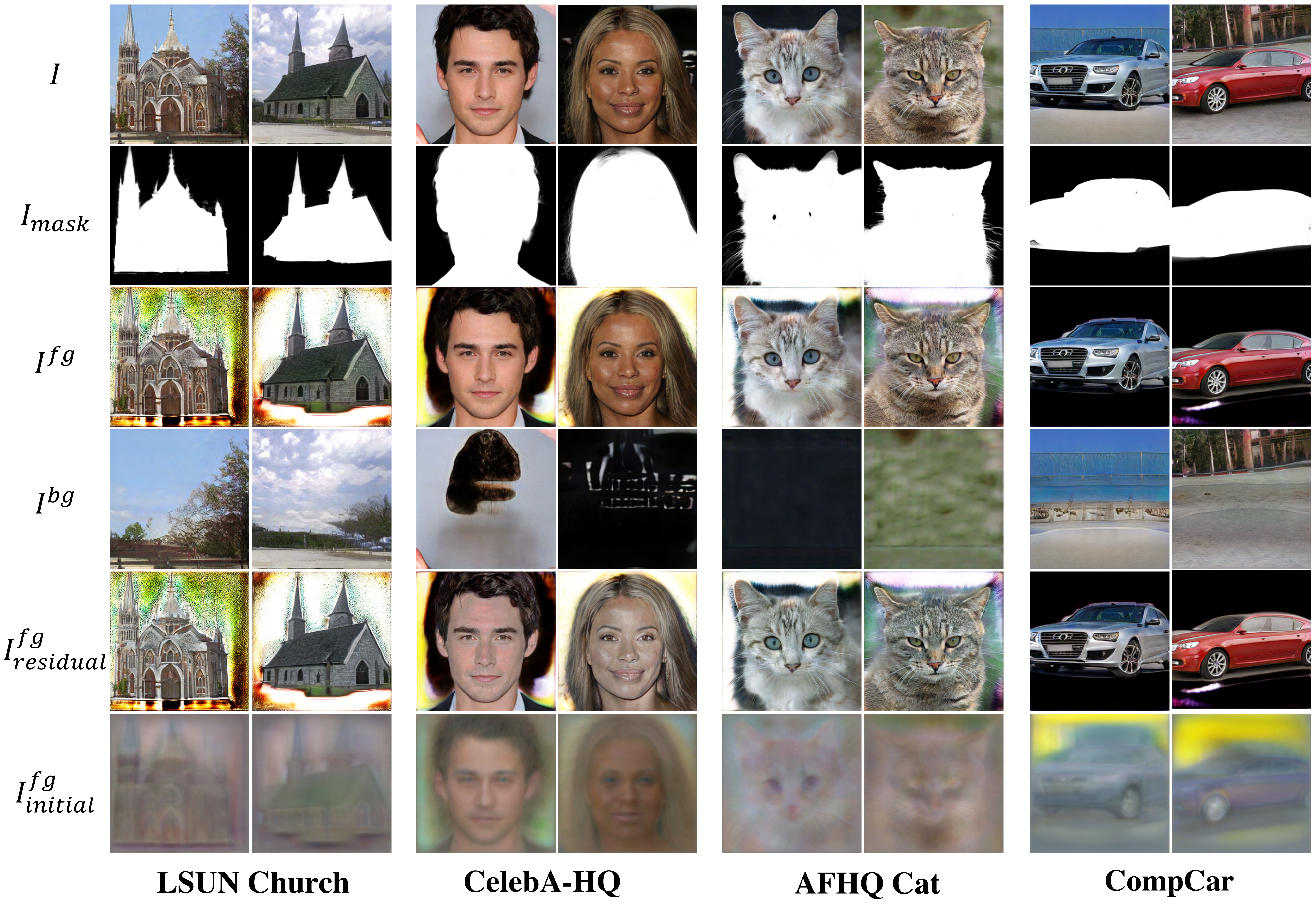}
    \caption{\textbf{Comprehensive Outputs.} We show all of GIRAFFE HD's intermediate and final output images for different datasets.}
    \label{fig:fg-bg_disen}
    \vspace{-0.1in}
\end{figure}

In Fig.~\ref{fig:3dcontrol}, we qualitatively demonstrate that our method preserves all of GIRAFFE's controllable features. For vertical translation, note how our position sharing enables the ground to move up with the car.  Also, compared to GIRAFFE, our background remains more consistent when the foreground changes, as shown in our lower MBS in Table~\ref{tab:mbs} (qualitative comparisons are in Figs.~\ref{fig:teaser} and~\ref{fig:giraffesamples} and supp).  This is due to our explicit separation of foreground and background generation. For objects that rest on a ground surface (e.g., cars), our model will also include the object's shadow as part of the foreground (see Fig.~\ref{fig:fg-bg_disen} for examples), which is the ideal behavior.  However, the DeepLabV3 model~\cite{chen2017rethinking} that is used to compute MBS does not segment the shadow as being part of the object, which is why our MBS is higher on CompCar than on FFHQ.  In Fig.~\ref{fig:fg-bg_disen}, we show comprehensive intermediate and final image generations.

\begin{table}[t!]
  \centering
  \footnotesize
  \begin{tabular}{@{}lcc@{}}
    \toprule
             & FFHQ & CompCar   \\

    \midrule
    GIRAFFE~\cite{Niemeyer2020GIRAFFE}      & 99.15 & 88.89     \\
    Ours                                    & 15.02 & 22.88     \\
    \bottomrule  
  \end{tabular}
  \caption{\textbf{Foreground-Background Disentanglement.} We report the MBS score (↓) for all methods on FFHQ and CompCar at $256^2$.}
  \label{tab:mbs}
\end{table}

\begin{table}[t!]
  \centering
  \footnotesize
  \begin{tabular}{@{}lc@{}}
    \toprule
                & CompCar $256^2$   \\
    \midrule
    w/o position sharing     & 10.89 \\  
    w/o environment sharing  & 11.55 \\
    full                     & 7.22  \\
    \bottomrule
  \end{tabular}
  \caption{\textbf{Ablation: Removing Position/Environment Sharing.} `full' denotes the full GIRAFFE HD model. We report FID (↓).}
  \label{tab:fgbgconsistency}
\end{table}

\begin{figure}[t!]
    \centering
    \includegraphics[width=0.48\textwidth]{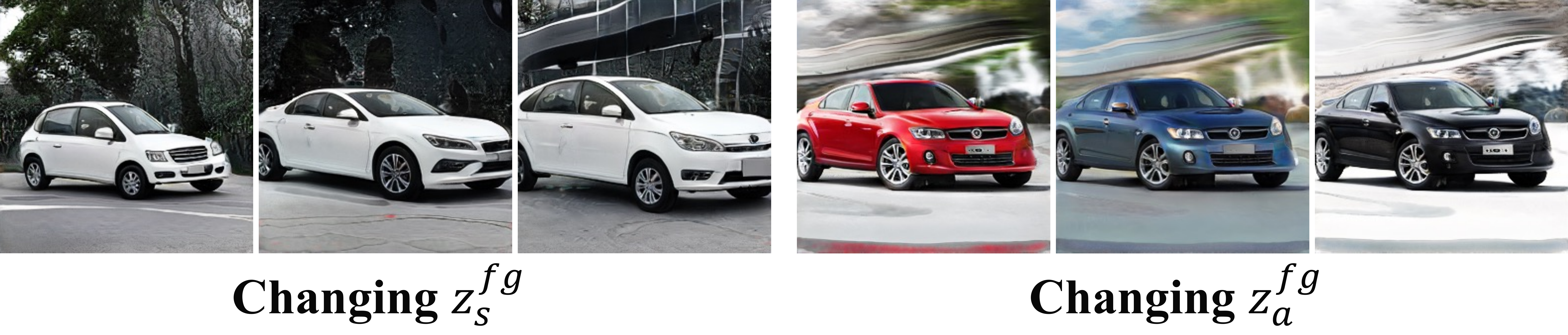}
    \caption{\textbf{Ablation: Single Style-based Renderer Baseline.} Notice how the foreground and background are entangled.}
    \label{fig:1stg_ablation}
\end{figure}

\subsection{Ablation studies}

\paragraph{Importance of two-stage.}
The most naive method for improving GIRAFFE's image quality is to simply replace GIRAFFE's neural renderer with a single style-based renderer. For this baseline, during training, we  use style mixing of $z_{s}^{fg}$ and $z_{a}^{fg}$ as latent codes to the renderer. During evaluation, we still inject $z_{s}^{fg}$ into the earlier levels and $z_{a}^{fg}$ into the later levels to ensure disentanglement between fine-grained shape and appearance. However, we observe that this single renderer baseline loses foreground-background disentanglement (Fig.~\ref{fig:1stg_ablation}). Even though the car's color remains the same when changing its shape, the background's shape changes as well. Similarly, even though the car's shape remains the same when changing its color, the background's color also changes. Even though the foreground and background are disentangled at the 3D feature level, since a single style-based renderer cannot separately control foreground and background, the foreground-background disentanglement is lost in the final 2D image.

\vspace{-10pt}
\paragraph{Importance of foreground-background consistency enforcement.}
Table~\ref{tab:fgbgconsistency} shows that removing either position or environment sharing hurts the model's FID, as final images whose foreground-background combinations are incompatible in geometry/photometry can be generated.

\begin{figure}[t!]
    \centering
    \includegraphics[width=0.34\textwidth]{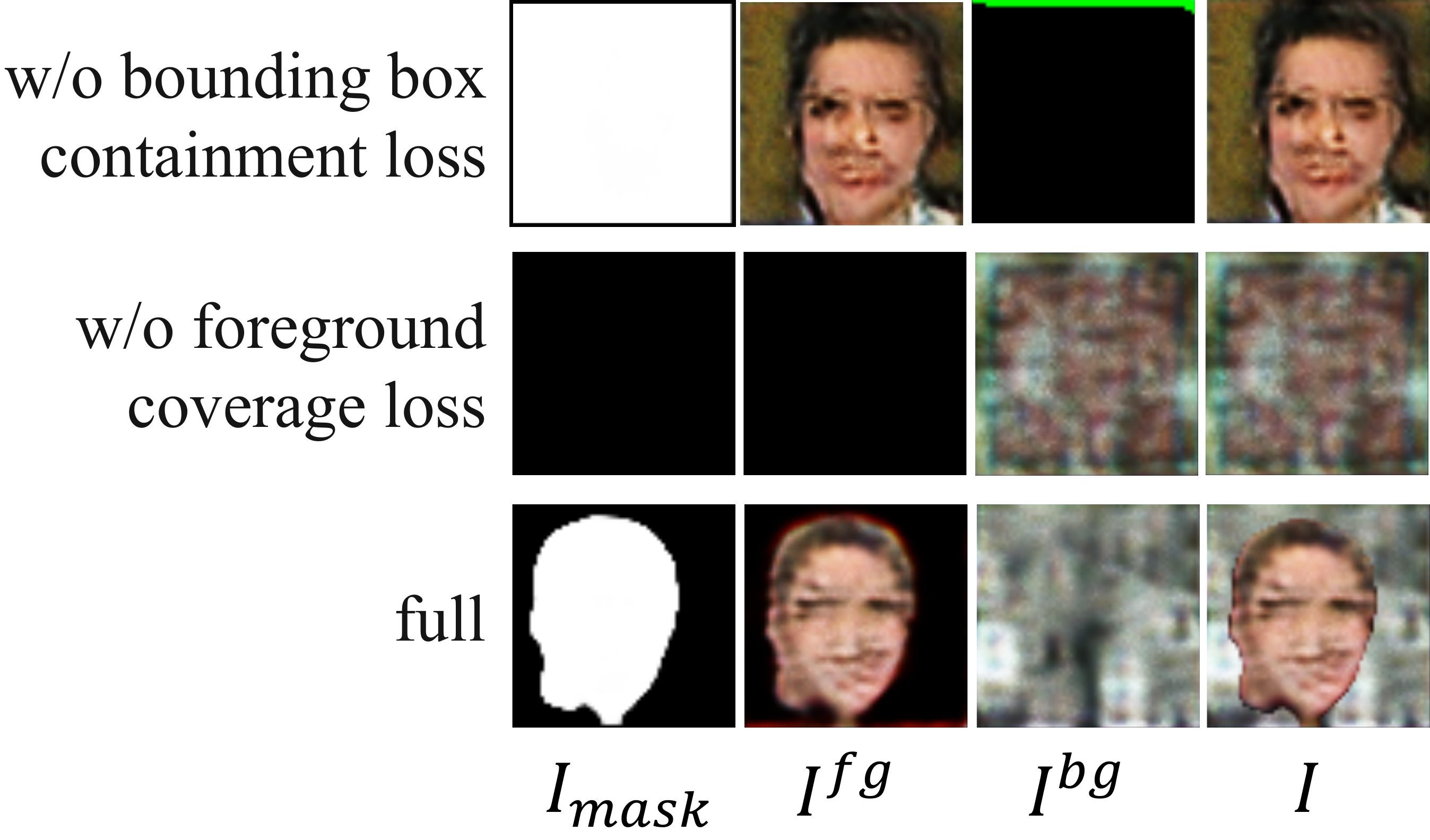}
    \caption{\textbf{Ablation: Removing Auxiliary Losses.} Even after just 4000 training iterations, the baselines that lack bounding box containment or foreground coverage loss generate all 1 or 0 masks.}
    \label{fig:loss_ablation}
\end{figure}

\vspace{-10pt}
\paragraph{Importance of auxiliary losses.}
In Fig.~\ref{fig:loss_ablation}, we show our model's $64^2$ renderings on FFHQ after 4000 training iterations, in three configurations. Without the bounding box containment loss the foreground branch generates the entire image, and without the foreground coverage loss the background generates the entire image. Hence these two losses are critical for foreground-background disentanglement.

\section{Discussion and Conclusion}

We proposed GIRAFFE HD, a high-resolution 3D-aware generative model that inherits all of GIRAFFE's~\cite{Niemeyer2020GIRAFFE} 3D controllable features while generating high-quality, high-resolution images.

\vspace{-10pt}
\paragraph{Limitations.}
First, we notice our model sometimes lacks 3D consistency. For example, when trained from scratch on the CompCar dataset, our model struggles to perform full 360 rotation. Instead, some shape codes correspond to front facing cars while others correspond to back facing cars, and each can only perform 180 rotation, even though the underlying 3D model has rotated 360 degrees. However, when we initialize the 3D feature generator with the weights of a pretrained GIRAFFE 3D feature generator and continue training, the model is then able to perform full 360 rotation.

Second, our current model (as well as GIRAFFE) cannot handle cross-domain foreground-background correlations when training a single model on multiple categories (e.g., cats, dogs, wildlife). Although the generated images are still high quality, there can be incompatible foreground and background combinations. Training a single model that works well on multiple datasets would be an interesting avenue for future work. 

Third, our current architecture cannot handle ``interlocking'' object relationships, i.e., there exist some rays along which an object appears in front of another object and also some rays behind. Our model works by composing an image with 2D layers, and the layer masks (object masks) for the composition are generated by considering only the object itself and not other objects in the scene. This can generally hold for most real-world domains. However, when objects ``interlock'', the objects' masks need to take into account other objects' 3D geometries. We think that a module that renders object masks based on 3D occlusion reasoning could be a possible extension to address this problem.

\vspace{-10pt}
\paragraph{Broader Impact.}
There are many possible applications for controllabe image generation including those in the entertainment and design industry.  For example, it could enhance the productivity of designers by allowing them to use such tools to control each object in the scene independently when creating new visual content.  Since our approach does not require supervision apart from having a collection of images of the same object, it is easily scalable to many different categories. However, there could also be potential misuses, such as creating fake content to fool law enforcement or to spread misinformation on social media. Recent work on models that can detect fake images (e.g.,~\cite{Wang2020CVPR}) could potentially be useful to prevent such unethical applications.

%-------------------------------------------------------------------------
\vspace{-10pt}
\paragraph{Acknowledgements.}
This work was supported in part by a Sony Focused Research Award and
NSF CAREER IIS-2150012.  We thank the anonymous reviewers for their constructive comments.

%%%%%%%%% REFERENCES
{\small
\bibliographystyle{ieee_fullname}
\bibliography{egbib}
}

\clearpage

\appendix
\paragraph{\Large Appendix}
 
\section{Full Loss Expression}
% \label{sec:intro}

For a given generator-discriminator pair $\{G,D\}$, the overall objective function can be formalized as
\begin{equation}
  \begin{aligned}
  &L(G,D)= \\
  &\mathbb{E}_{z_a^{k}, z_s^{k} \sim \mathcal{N}, \xi \sim p_\xi, T^k \sim p_T}[f(D(G(\{z_a^{k}, z_s^{k}, T^k\}_k, \xi)))]\\
  &+ \mathbb{E}_{I \sim p_D}[f(-D(I))- \frac{\lambda}{2} \Vert \nabla D(I) \Vert ^2]\\
  &+ \beta_1 L_{bbox} + \beta_2 L_{cvg} + \beta_3 L_{bin}
  \end{aligned}
%   \label{eq:pos-encoding}
\end{equation}
where $f(t) = - \log(1 + \exp(-t))$, $\lambda = 10$, $p_D$ indicates the data distribution, and $\beta_1, \beta_2, \beta_3$ are dataset specific. $L_{bbox}$,  $L_{cvg}$, and $L_{bin}$ are as defined in the main paper.

\section{Mutual Background Similarity (MBS) Details}
% \label{sec:intro}

We denote the generator to evaluate as $G$, which takes as input randomly sampled foreground parameters $P_{fg}\sim p_{fg}$ and background parameters $P_{bg}\sim p_{bg}$ to generate an image $I$. We denote a pretrained semantic segmentation model DeepLabV3 ResNet101~\cite{chen2017rethinking} as $R$ which takes an image $I$ and outputs the semantic prediction map for $I$, which can then be converted into the background mask $M$. We compute the mutual background similarity (MBS) by first randomly sampling an image $I_1 = G(P_{fg_1}, P_{bg})$, then generating another image by sampling another $P_{fg_2}\sim p_{fg}$ while keeping $P_{bg}$ fixed, $I_2 = G(P_{fg_2}, P_{bg})$. Then we compute the background masks for the two images $M_1 = R(I_1), M_2 = R(I_2)$ and the mask for the two images' mutual background area can be computed as $M_{multbg} = M_1 \cdot M_2$. We define that a pixel's RGB value has changed if one or more channels of the pixel's RGB value has changed over some small threshold $\eta$. Then the total number of pixels inside the mutual background area whose RGB value has changed is computed as
 
\begin{equation}
N = {\sum}_{i\in M_{multbg}} \delta
%   \label{eq:pos-encoding}
\end{equation}

\[
\text{where } \delta=
\begin{cases}
    0,& \text{if } \eta > |I_1[i][c]-I_2[i][c]| , c\in \{R, G, B\}\\
    1,              & \text{otherwise}
\end{cases}
\]
The image is normalized to [0,1] before feeding into $R$, and $\eta$ is set to be $\frac{1}{255}$. Then the MBS for image pair $\{I_1, I_2\}$ is

\begin{equation}
MBS = \frac{N}{|M_{multbg}|} \times 100
%   \label{eq:pos-encoding}
\end{equation}

\begin{figure}[h!]
    \centering
    \includegraphics[width=0.45\textwidth]{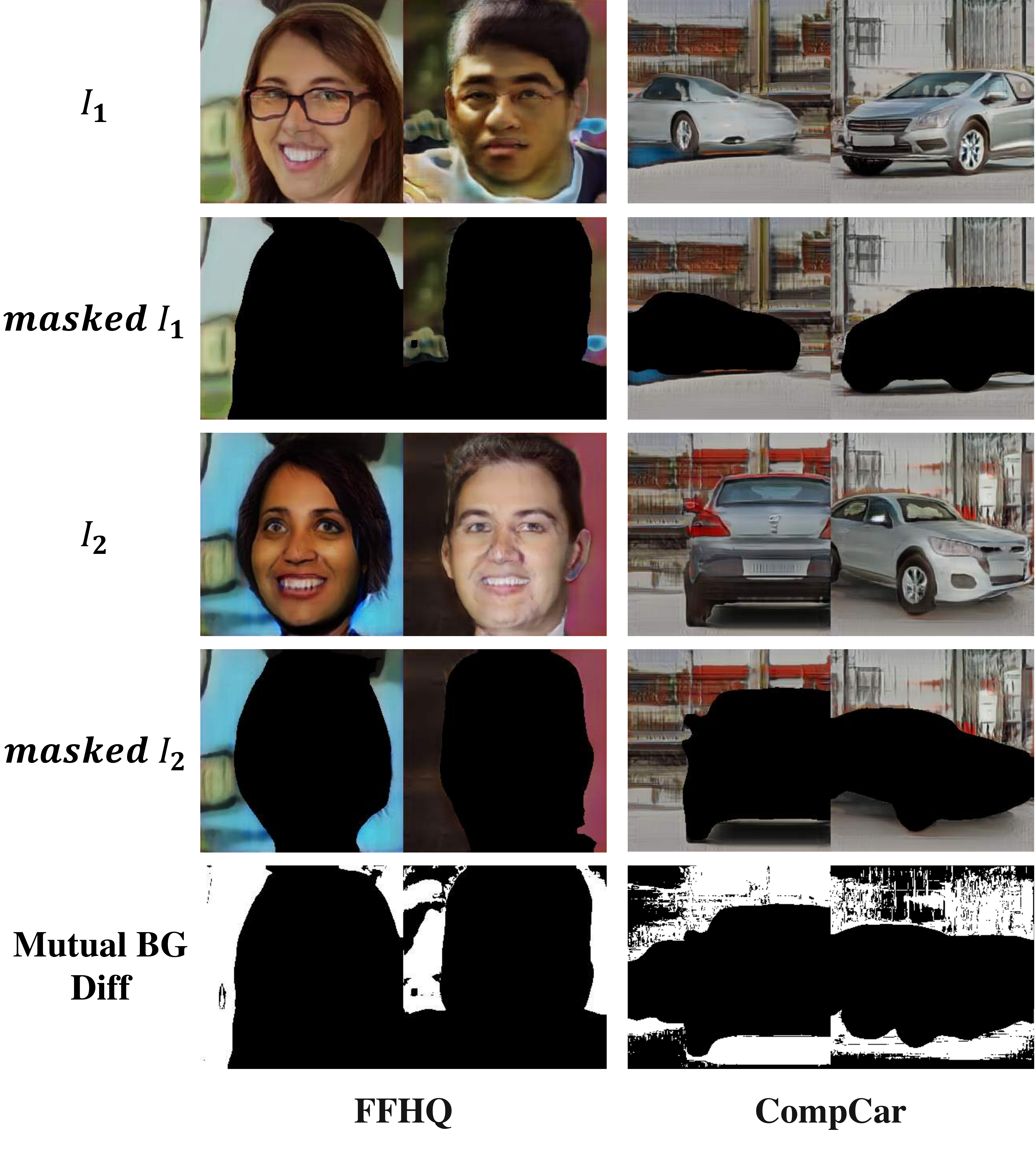}
    \caption{\textbf{GIRAFFE MBS Calculation.} DeepLabV3 background segmentations and mutual background differences (white pixels) used for computing MBS on GIRAFFE samples.}
    \label{fig:grf_mbs}
\end{figure}

\begin{figure}[h!]
    \centering
    \includegraphics[width=0.45\textwidth]{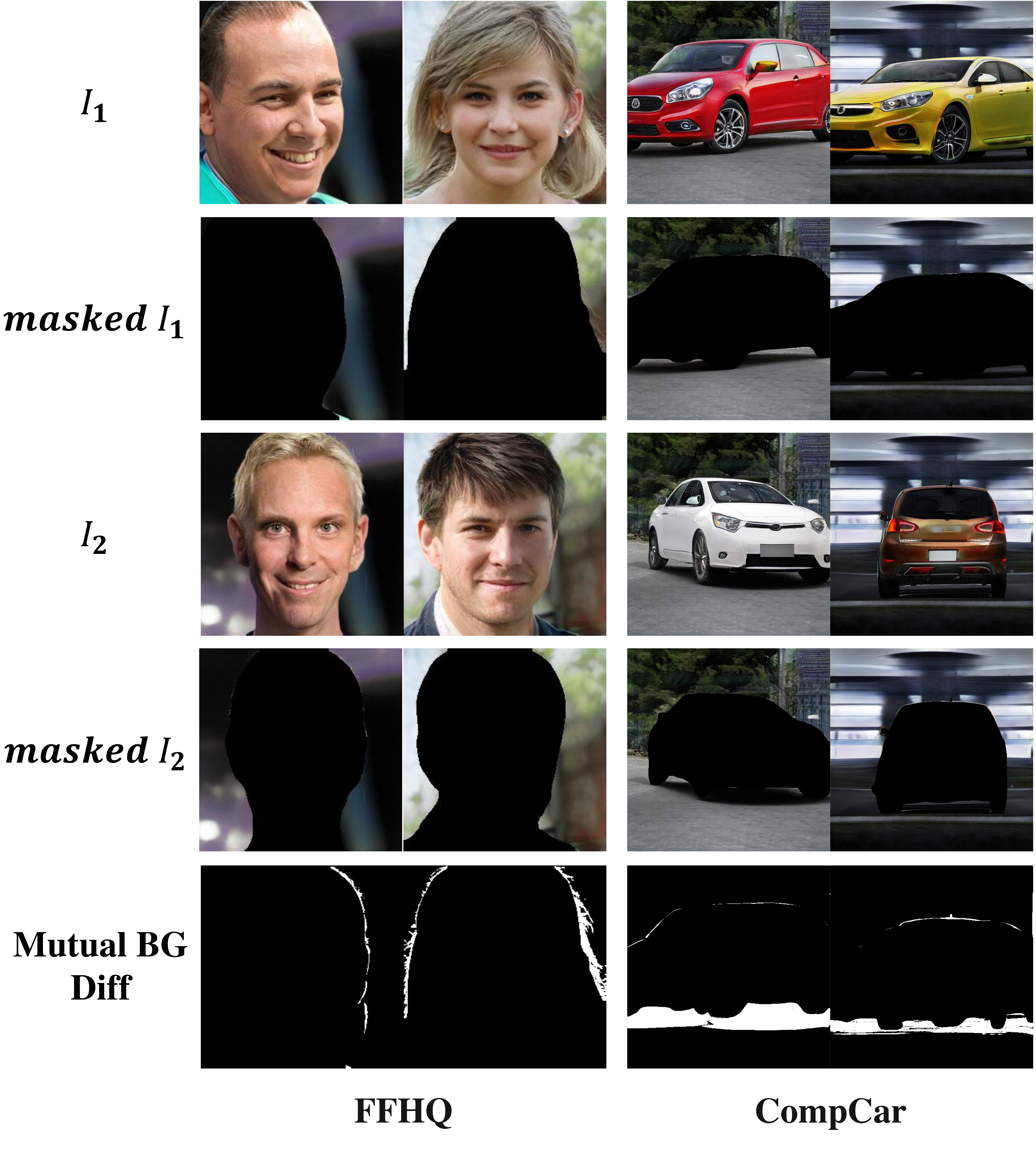}
    \caption{\textbf{GIRAFFE HD (ours) MBS Calculation.} DeepLabV3 background segmentations and mutual background differences (white pixels) used for computing MBS on our GIRAFFE HD samples.}
    \label{fig:grfhd_mbs}
\end{figure}

In Figures~\ref{fig:grf_mbs} and~\ref{fig:grfhd_mbs}, we show the segmentations produced by DeepLabV3 ResNet101~\cite{chen2017rethinking} and the mutual background difference map for both GIRAFFE and GIRAFFE HD (ours) on FFHQ~\cite{karras-cvpr2019} and CompCar~\cite{compcars} datasets. For GIRAFFE HD on FFHQ, the mutual background difference mainly comes from the imprecision of the segmentation (as DeepLabV3 cannot properly segment thin, floating hair). For GIRAFFE HD on CompCar, the mutual background difference mainly comes from the segmentor not including the car's shadow as part of the foreground.

\section{Dataset Details}
% \label{sec:intro}
\paragraph{Dataset parameters.}
We report the dataset-dependent camera
elevation angle and valid object transformation parameters used for all the datasets in Table~\ref{tab:params}. We use the same dataset parameters as GIRAFFE for CompCar, FFHQ, LSUN Church and CelebA-HQ datasets (except for CompCar's vertical translation). Since GIRAFFE was not evaluated on AFHQ Cat, we use the same dataset parameters GIRAFFE uses for Cats~\cite{cathead}.

\section{Additional Qualitative Results}
\begin{figure*}[t!]
    \centering
    \includegraphics[width=.99\textwidth]{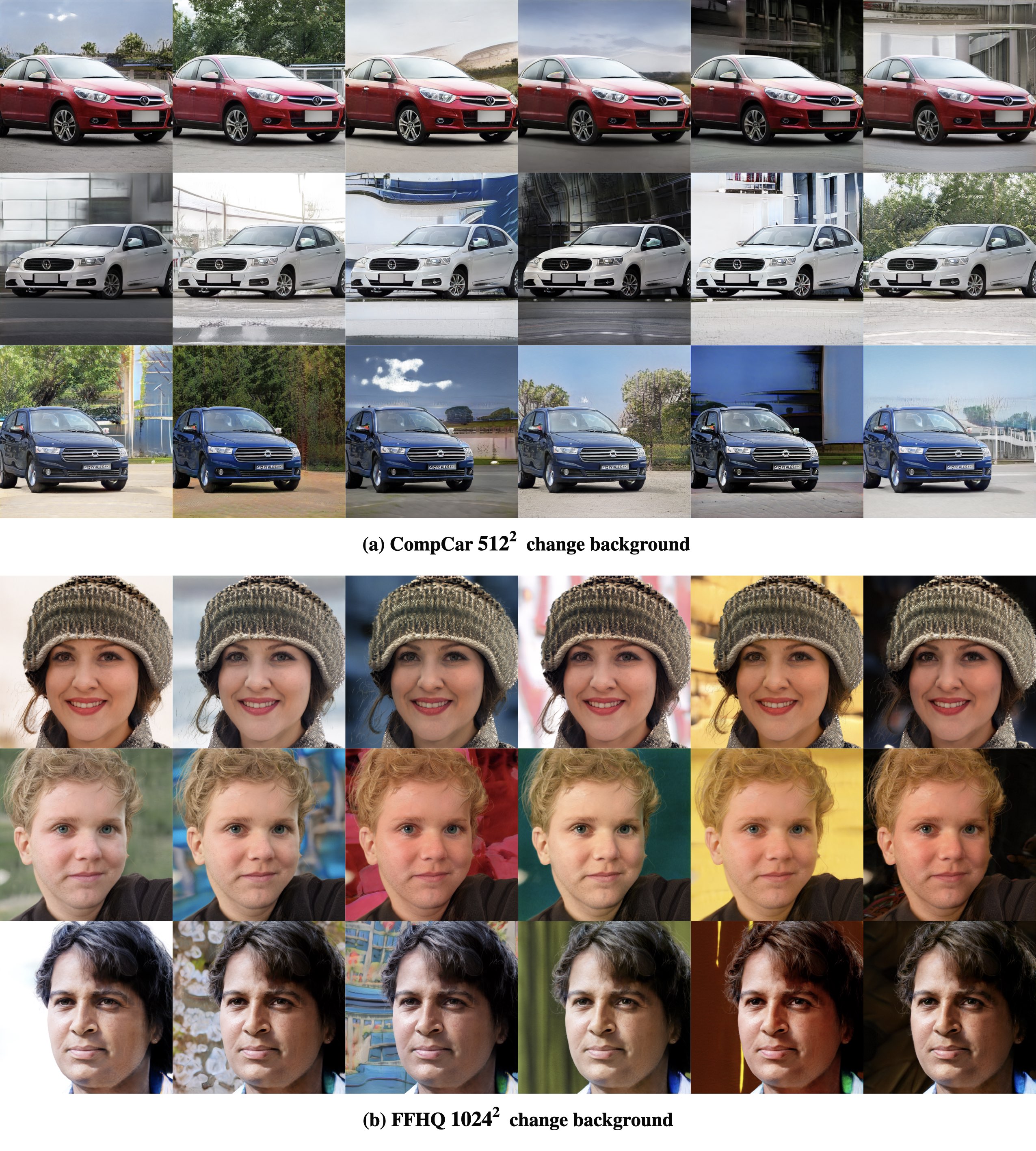}
    \caption{\textbf{Controllable Image Synthesis.} Changing background results on CompCar~\cite{compcars} and FFHQ~\cite{karras-cvpr2019}.  Notice how the appearance of the foreground adapts to the changing background.}
    \label{fig:change_bg1}
\end{figure*}

\begin{figure*}[t!]
    \centering
    \includegraphics[width=.99\textwidth]{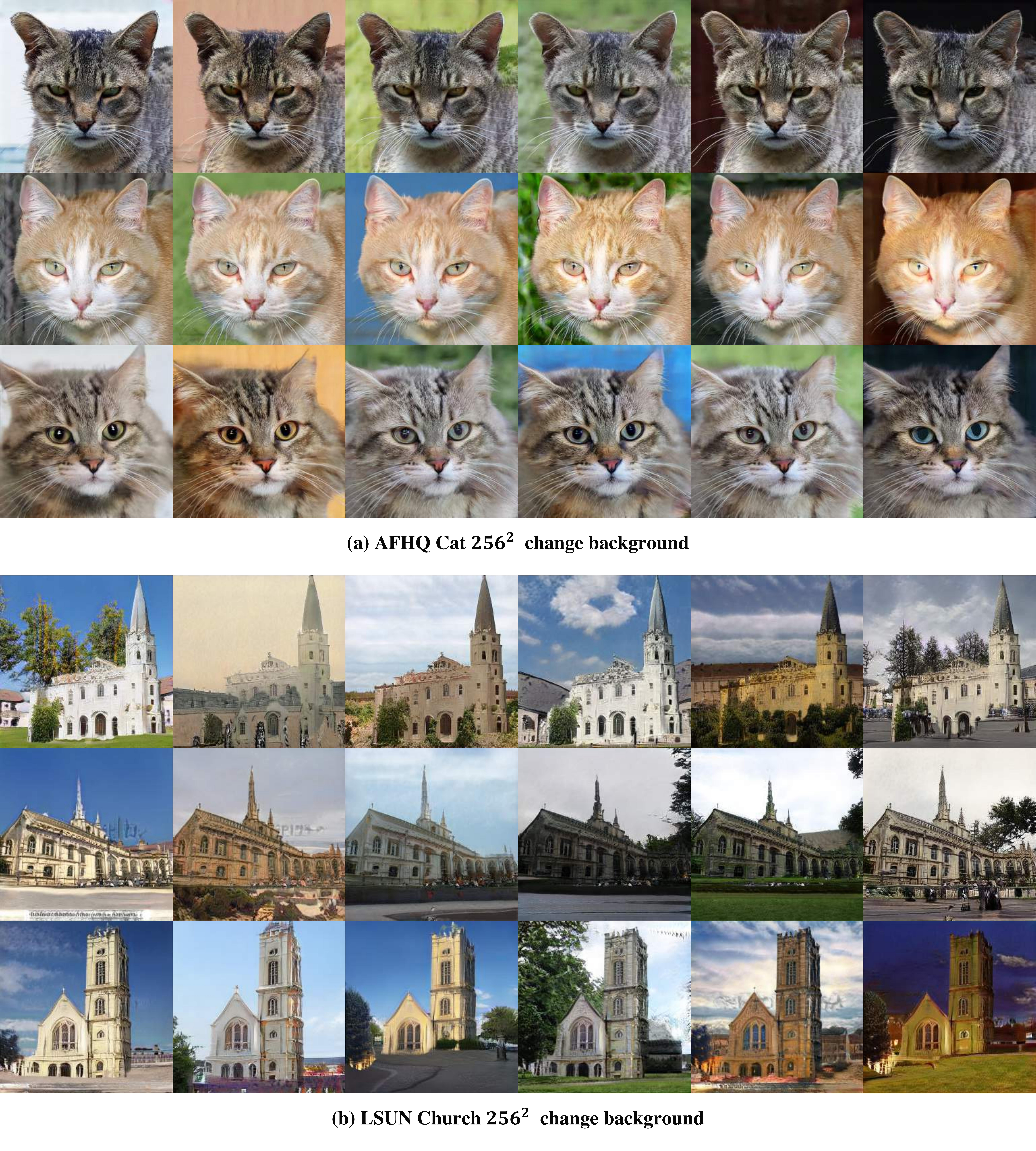}
    \caption{\textbf{Controllable Image Synthesis.} Changing background results on AFHQ Cat~\cite{Starganv2} and LSUN Church~\cite{Yu2015LSUNCO}. Notice how the appearance of the foreground adapts to the changing background. We also observe that for datasets where the foreground object does not have great variation in appearance (e.g., LSUN Church), the  refine foreground renderer tends to take more control over the final foreground object's appearance than the initial foreground renderer. In such cases, making changes to the background tends to change the foreground appearance more.}
    \label{fig:change_bg2}
\end{figure*}

\begin{figure*}[t!]
    \centering
    \includegraphics[width=.99\textwidth]{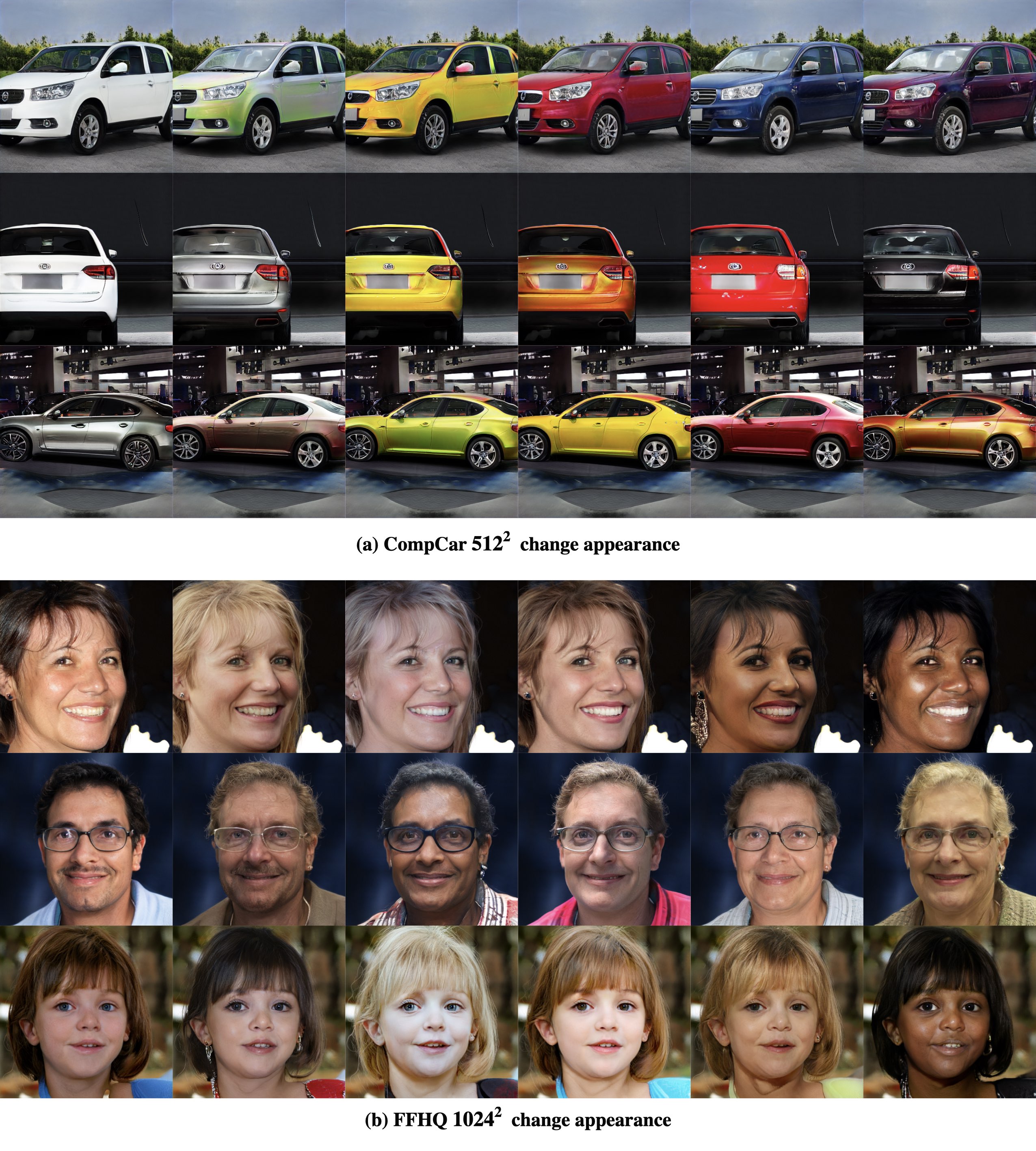}
    \caption{\textbf{Controllable Image Synthesis.} Changing appearance results on CompCar~\cite{compcars} and FFHQ~\cite{karras-cvpr2019}.}
    \label{fig:change_app1}
\end{figure*}

\begin{figure*}[t!]
    \centering
    \includegraphics[width=.99\textwidth]{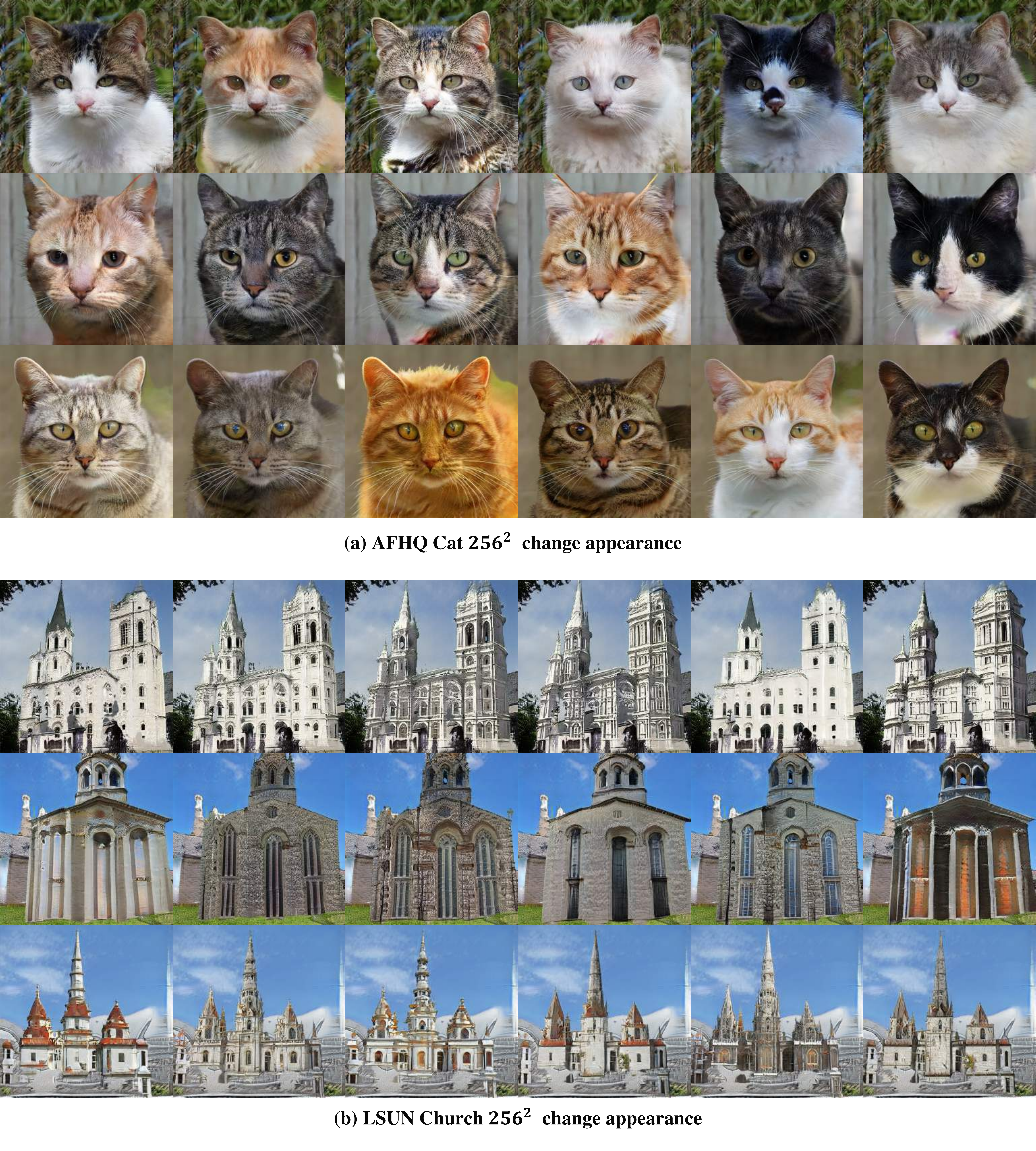}
    \caption{\textbf{Controllable Image Synthesis.} Changing appearance results on AFHQ Cat~\cite{Starganv2} and LSUN Church~\cite{Yu2015LSUNCO}. As mentioned previously, for datasets where the foreground object does not have great variation in appearance (e.g., LSUN Church), the  refine foreground renderer tends to take more control over the final foreground object's appearance than the initial foreground renderer. In such cases, making changes to the foreground appearance code tends to have relatively less effect on the appearance of the foreground object.}
    \label{fig:change_app2}
\end{figure*}

\begin{figure*}[t!]
    \centering
    \includegraphics[width=.99\textwidth]{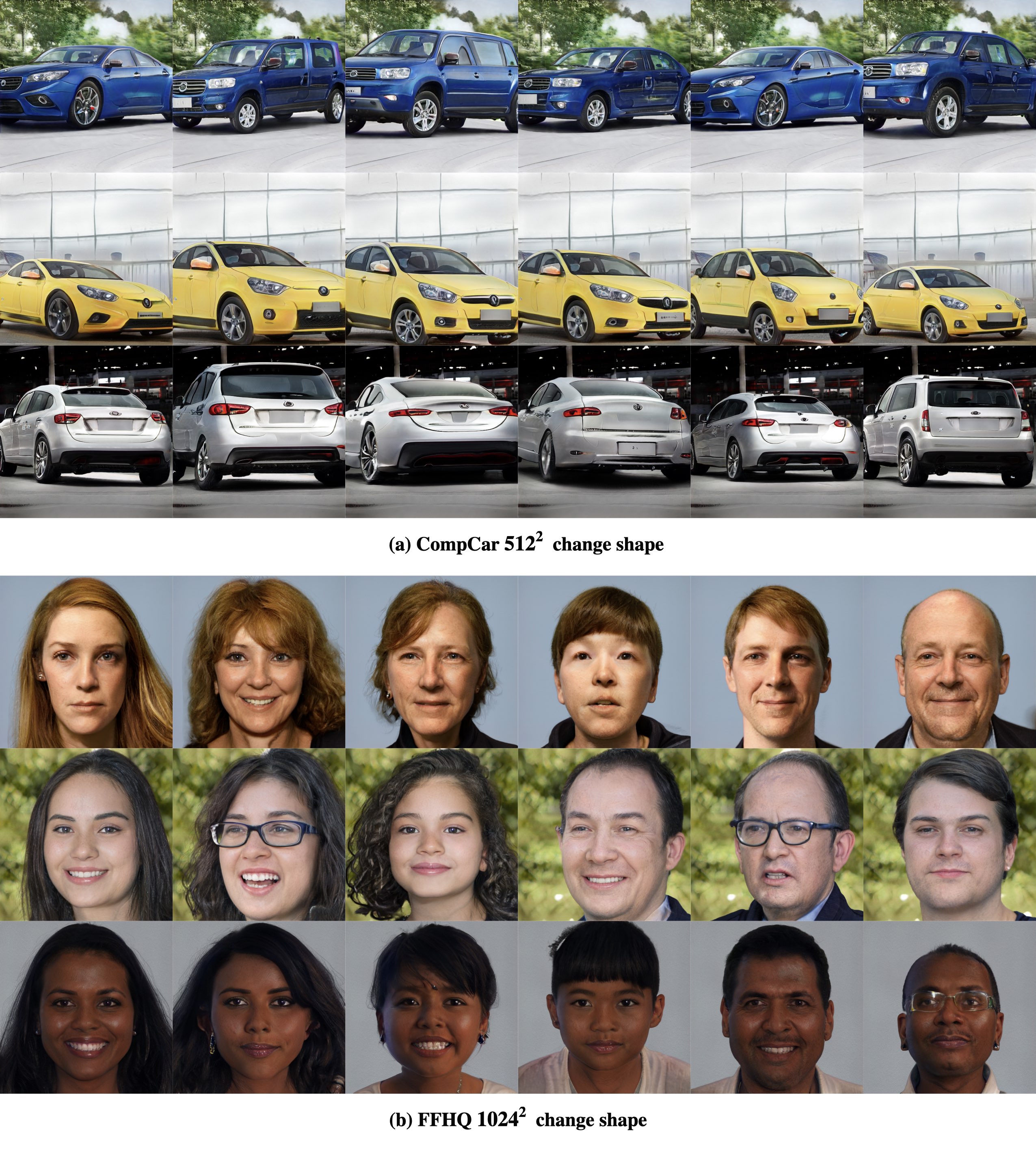}
    \caption{\textbf{Controllable Image Synthesis.} Changing shape results on CompCar~\cite{compcars} and FFHQ~\cite{karras-cvpr2019}.}
    \label{fig:change_shape1}
\end{figure*}

\begin{figure*}[t!]
    \centering
    \includegraphics[width=.99\textwidth]{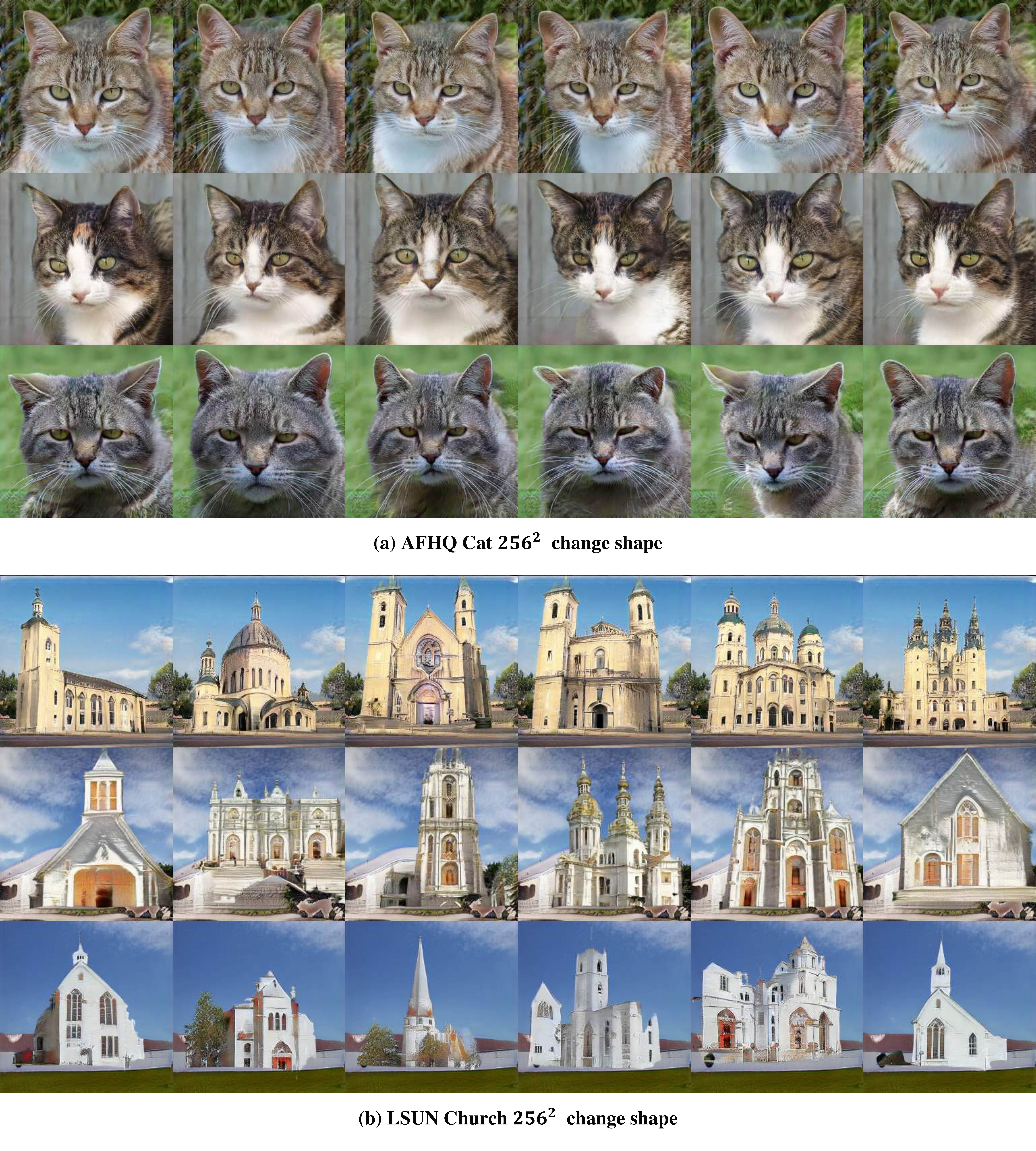}
    \caption{\textbf{Controllable Image Synthesis.} Changing shape results on AFHQ Cat~\cite{Starganv2} and LSUN Church~\cite{Yu2015LSUNCO}.}
    \label{fig:change_shape2}
\end{figure*}

\begin{figure*}[t!]
    \centering
    \includegraphics[width=.99\textwidth]{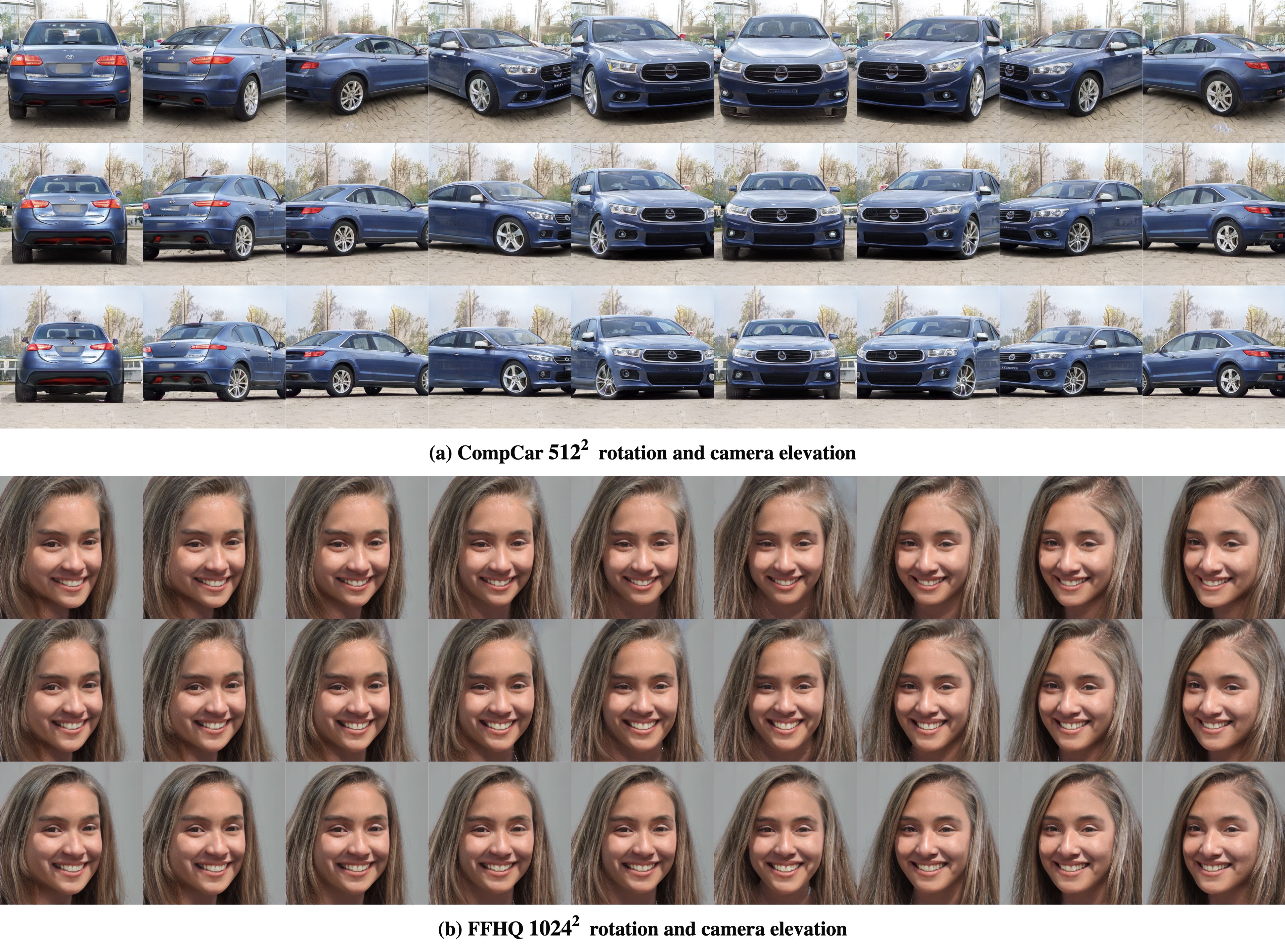}
    \caption{\textbf{Controllable Image Synthesis.} Changing rotation and camera elevation results on CompCar~\cite{compcars} and FFHQ~\cite{karras-cvpr2019}.}
    \label{fig:change_rot1}
\end{figure*}

\begin{figure*}[t!]
    \centering
    \includegraphics[width=.99\textwidth]{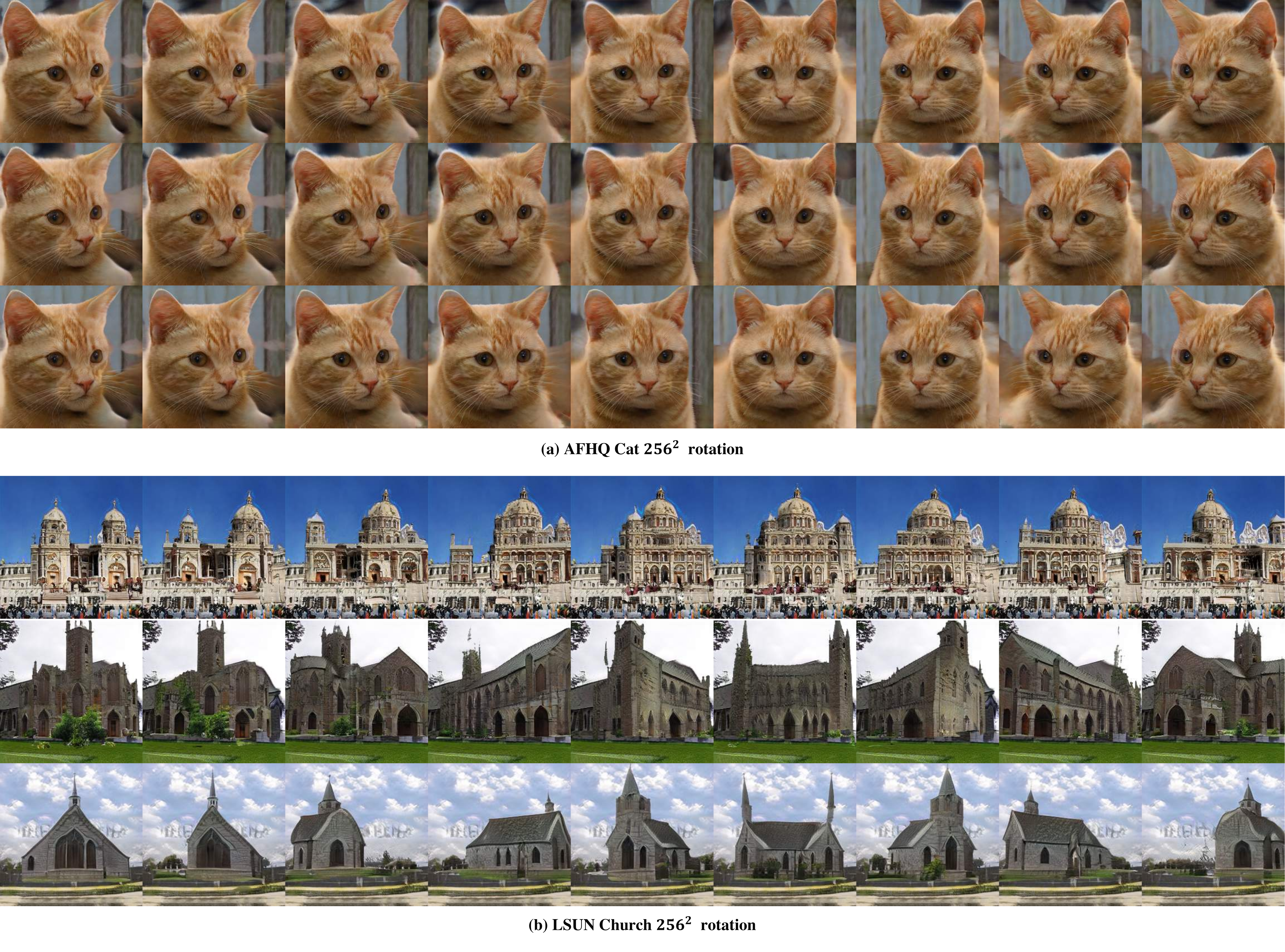}
    \caption{\textbf{Controllable Image Synthesis.} Changing rotation and camera elevation results on AFHQ Cat~\cite{Starganv2} and changing rotation results on LSUN Church~\cite{Yu2015LSUNCO} (the model is trained with a fixed camera elevation on the LSUN Church dataset). We observe that changing the camera elevation has little effect on the AFHQ Cat results. We attribute this to its small dataset size.}
    \label{fig:change_rot2}
\end{figure*}

\begin{figure*}[t!]
    \centering
    \includegraphics[width=.99\textwidth]{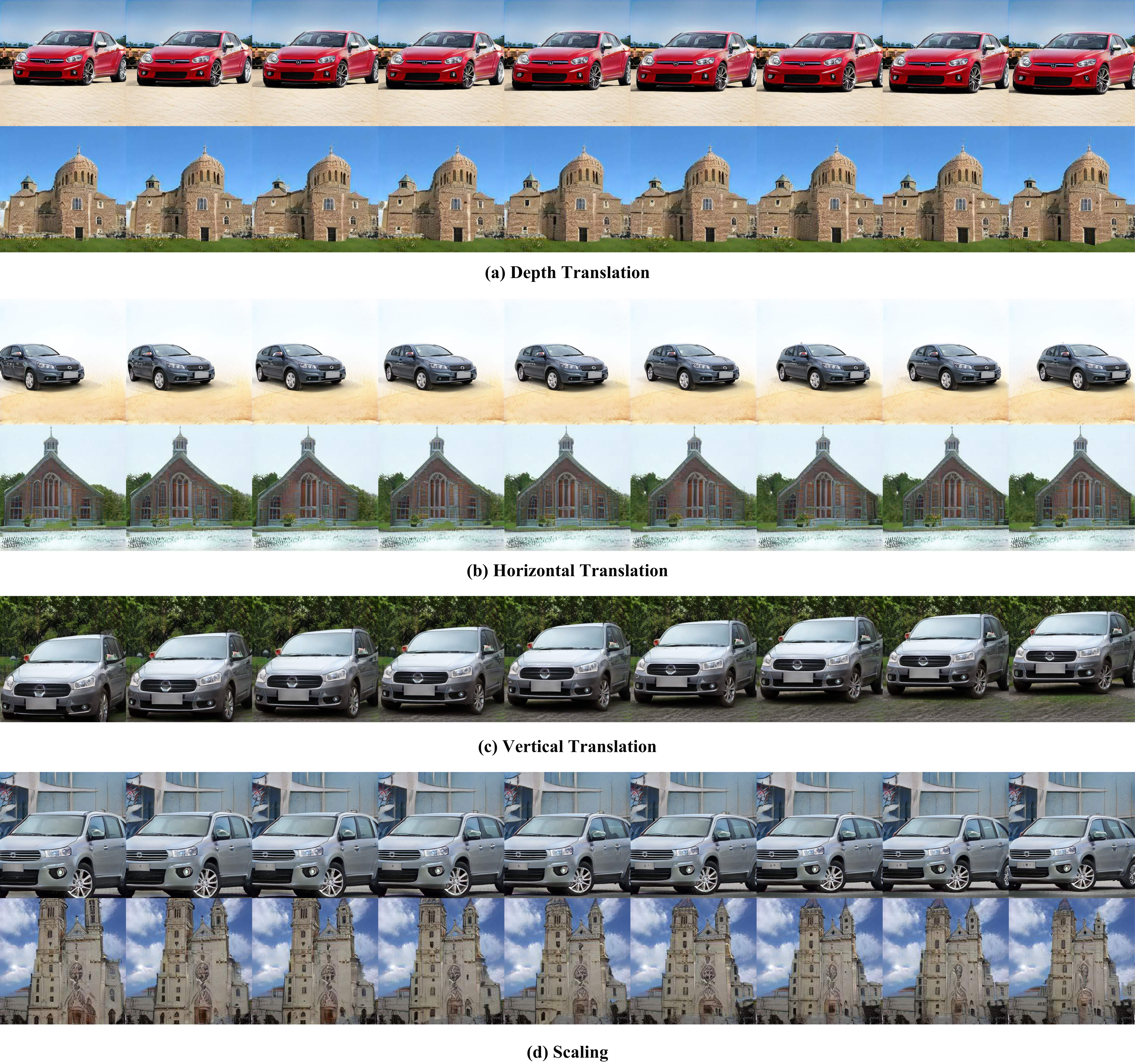}
    \caption{\textbf{Controllable Image Synthesis.} Translation and scaling results on CompCar~\cite{compcars} and LSUN Church~\cite{Yu2015LSUNCO}.}
    \label{fig:change_translation}
\end{figure*}

\begin{figure*}[t!]
    \centering
    \includegraphics[width=.99\textwidth]{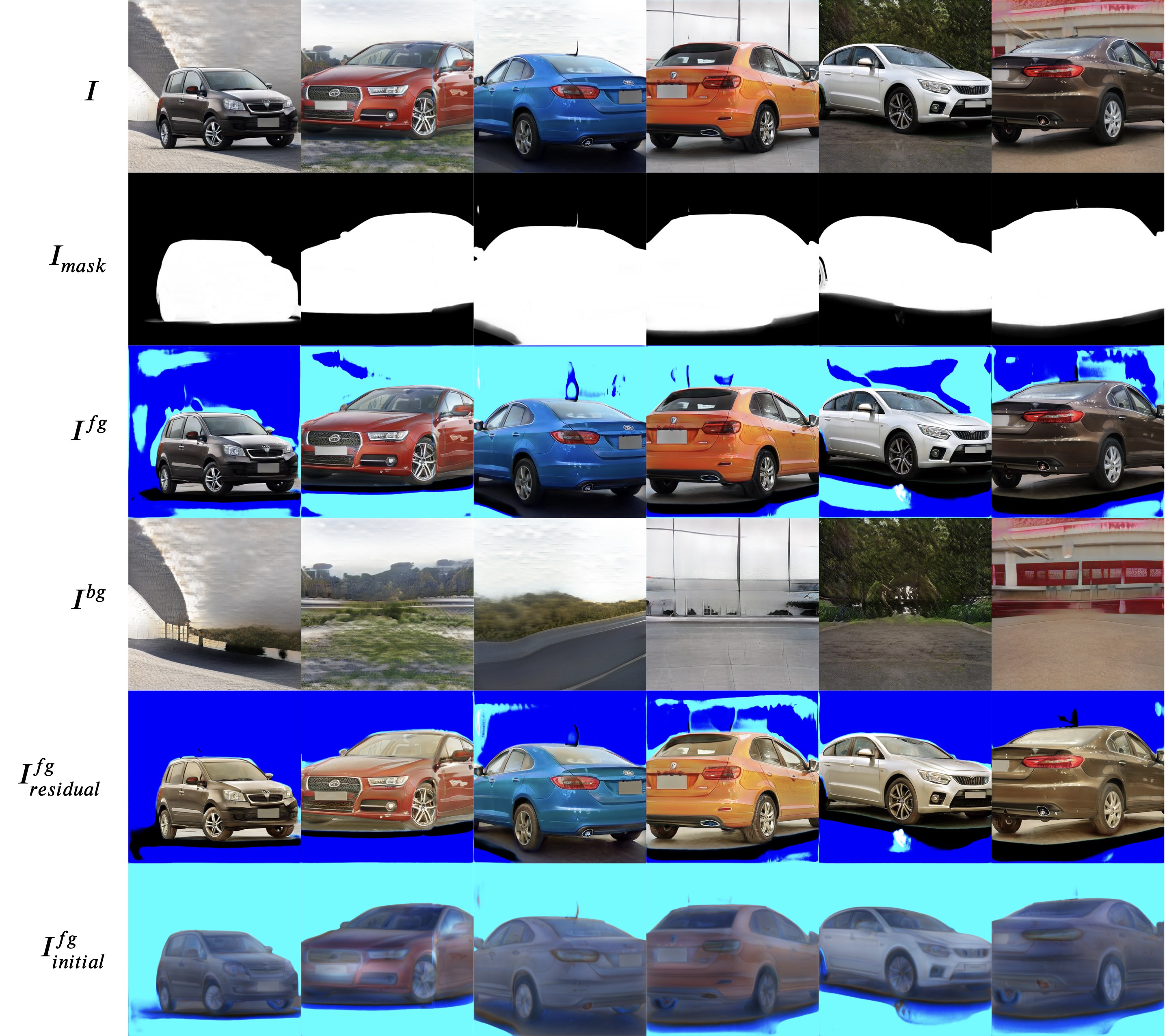}
    \caption{\textbf{Comprehensive Outputs.} Intermediate and final output images for CompCar~\cite{compcars} $512^2$.}
    \label{fig:comp1}
\end{figure*}

\begin{figure*}[t!]
    \centering
    \includegraphics[width=.99\textwidth]{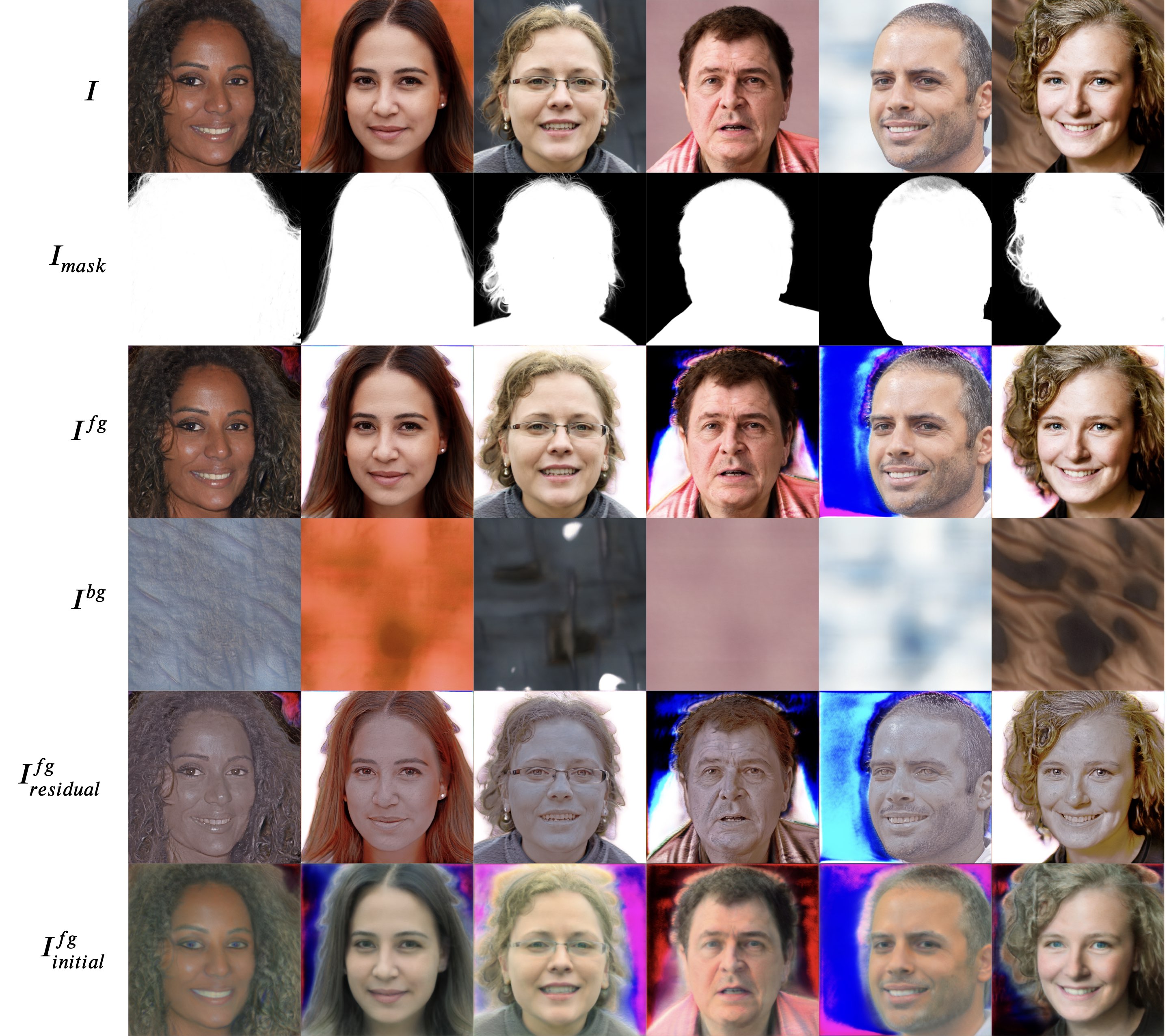}
    \caption{\textbf{Comprehensive Outputs.} Intermediate and final output images for FFHQ~\cite{karras-cvpr2019} $1024^2$.}
    \label{fig:comp2}
\end{figure*}

\begin{figure*}[t!]
    \centering
    \includegraphics[width=.99\textwidth]{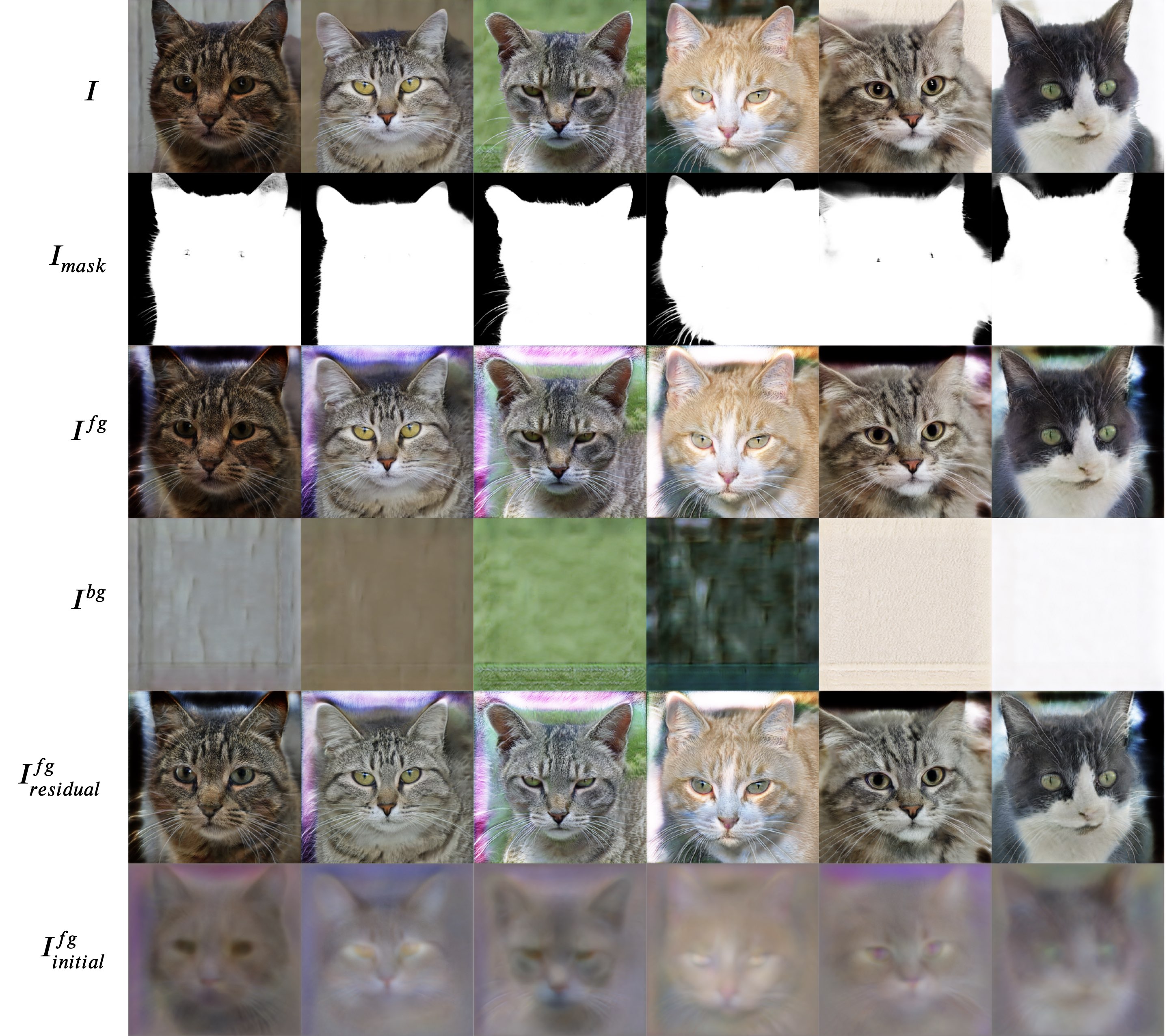}
    \caption{\textbf{Comprehensive Outputs.} Intermediate and final output images for AFHQ Cat~\cite{Starganv2} $256^2$.}
    \label{fig:comp3}
\end{figure*}

\begin{figure*}[t!]
    \centering
    \includegraphics[width=.99\textwidth]{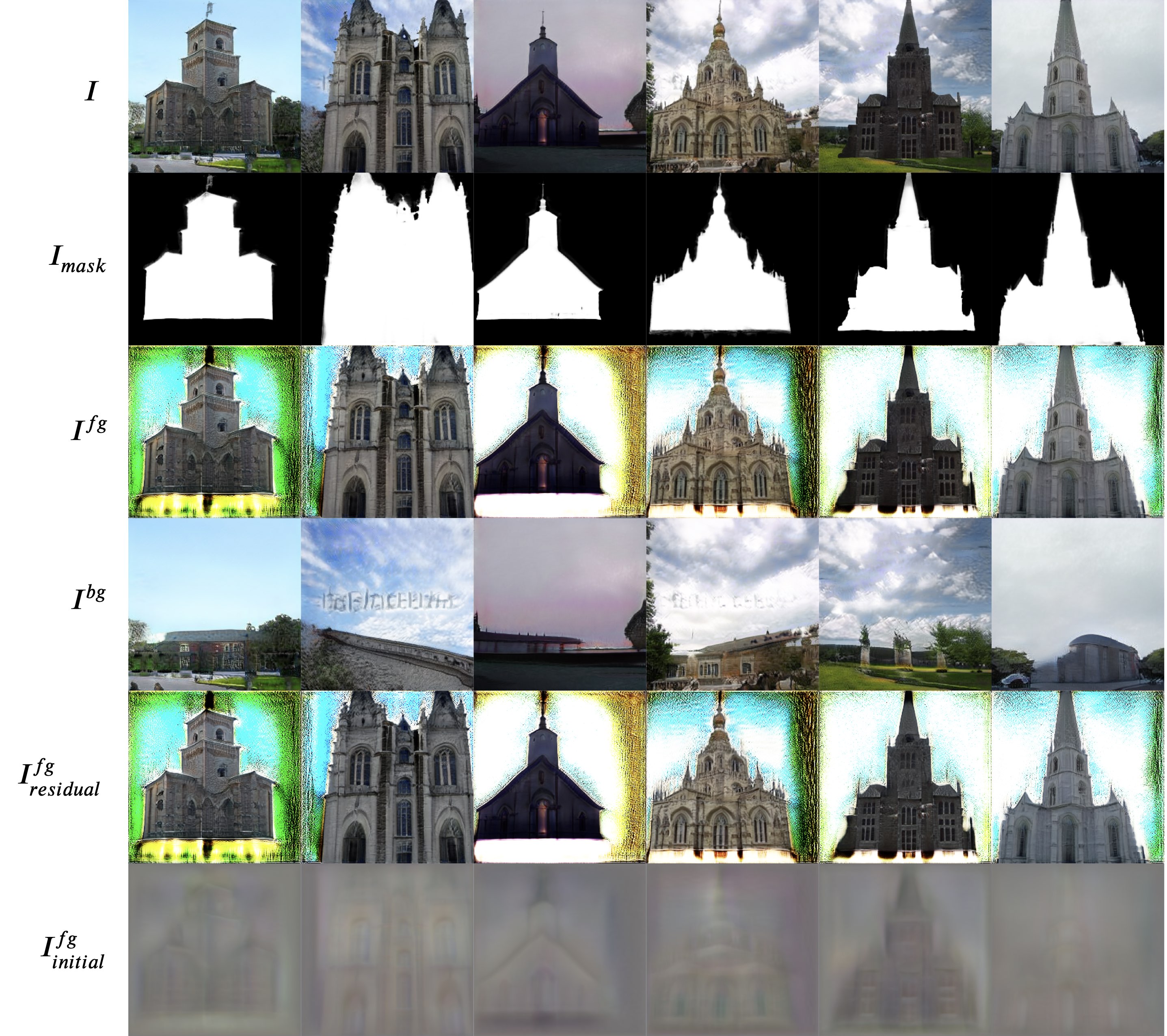}
    \caption{\textbf{Comprehensive Outputs.} Intermediate and final output images for LSUN Church~\cite{Yu2015LSUNCO} $256^2$.}
    \label{fig:comp4}
\end{figure*}

\begin{figure*}[t!]
    \centering
    \includegraphics[width=.99\textwidth]{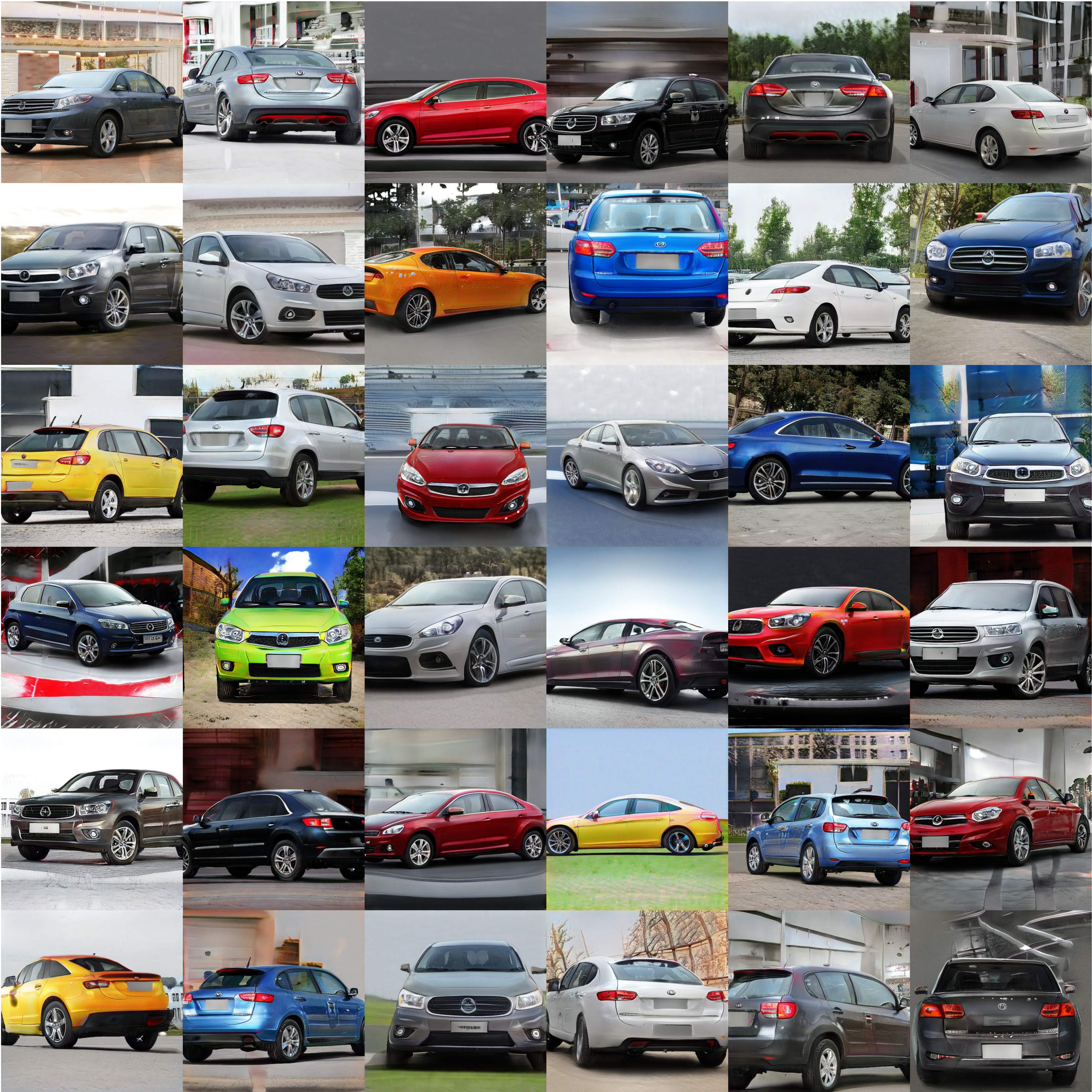}
    \caption{\textbf{Our samples.} GIRAFFE HD samples on CompCar~\cite{compcars} $512^2$ .}
    \label{fig:sample1}
\end{figure*}

\begin{figure*}[t!]
    \centering
    \includegraphics[width=.99\textwidth]{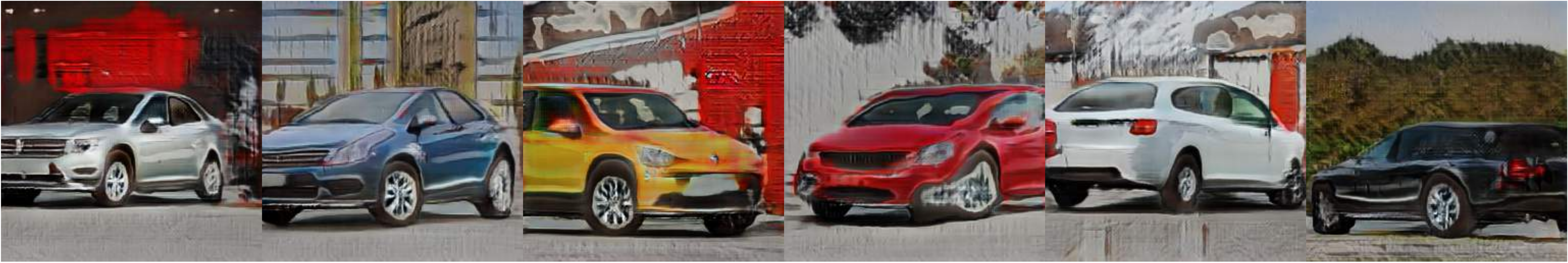}
    \caption{\textbf{GIRAFFE~\cite{Niemeyer2020GIRAFFE} samples.} GIRAFFE samples on CompCar $256^2$.}
    \label{fig:grf_sample1}
\end{figure*}

\begin{figure*}[t!]
    \centering
    \includegraphics[width=.99\textwidth]{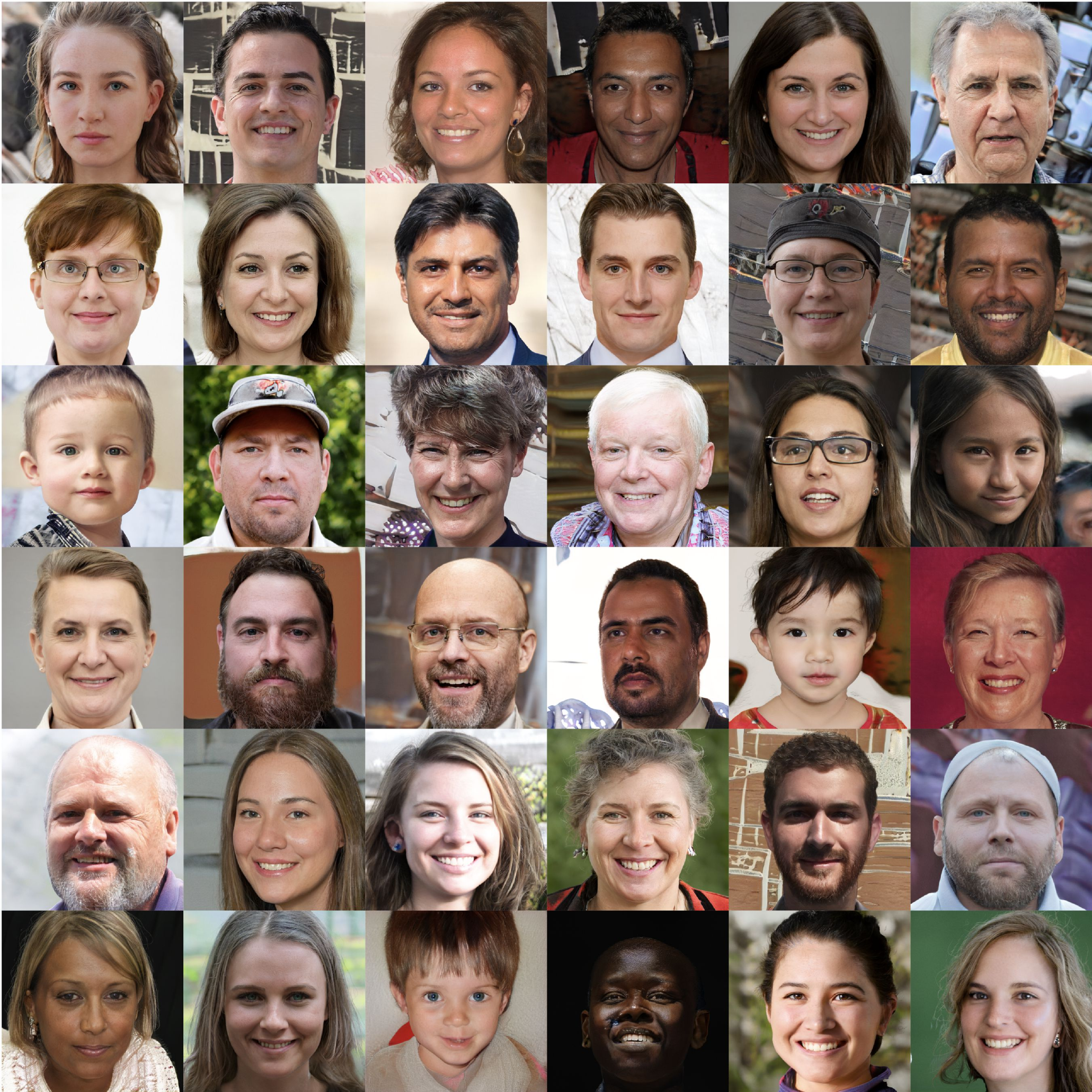}
    \caption{\textbf{Our samples.} GIRAFFE HD samples on FFHQ~\cite{karras-cvpr2019} $1024^2$.}
    \label{fig:sample2}
\end{figure*}

\begin{figure*}[t!]
    \centering
    \includegraphics[width=.99\textwidth]{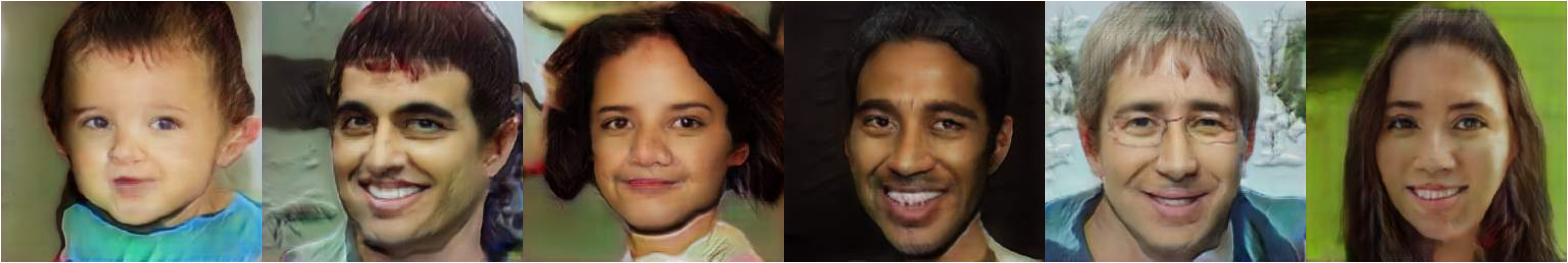}
    \caption{\textbf{GIRAFFE~\cite{Niemeyer2020GIRAFFE} samples.} GIRAFFE samples on FFHQ $256^2$.}
    \label{fig:grf_sample2}
\end{figure*}

\begin{figure*}[t!]
    \centering
    \includegraphics[width=.99\textwidth]{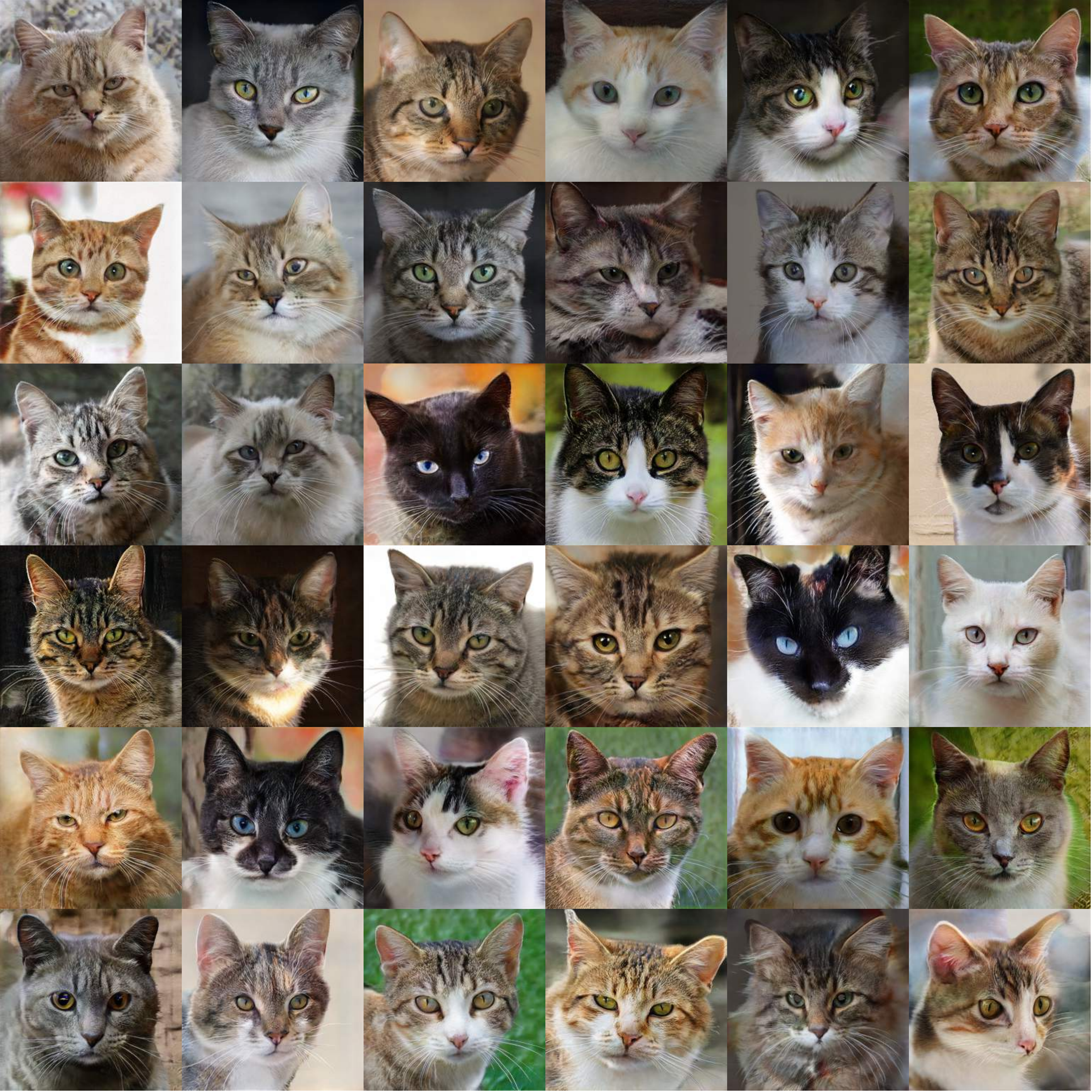}
    \caption{\textbf{Our samples.} GIRAFFE HD samples on AFHQ Cat~\cite{Starganv2} $256^2$.}
    \label{fig:sample3}
\end{figure*}

\begin{figure*}[t!]
    \centering
    \includegraphics[width=.99\textwidth]{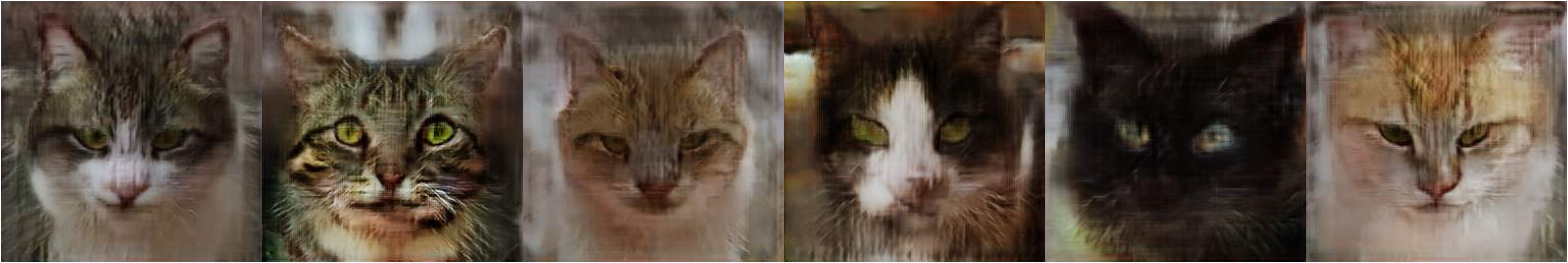}
    \caption{\textbf{GIRAFFE~\cite{Niemeyer2020GIRAFFE} samples.} GIRAFFE samples on AFHQ Cat $256^2$.}
    \label{fig:grf_sample4}
\end{figure*}

\begin{figure*}[t!]
    \centering
    \includegraphics[width=.99\textwidth]{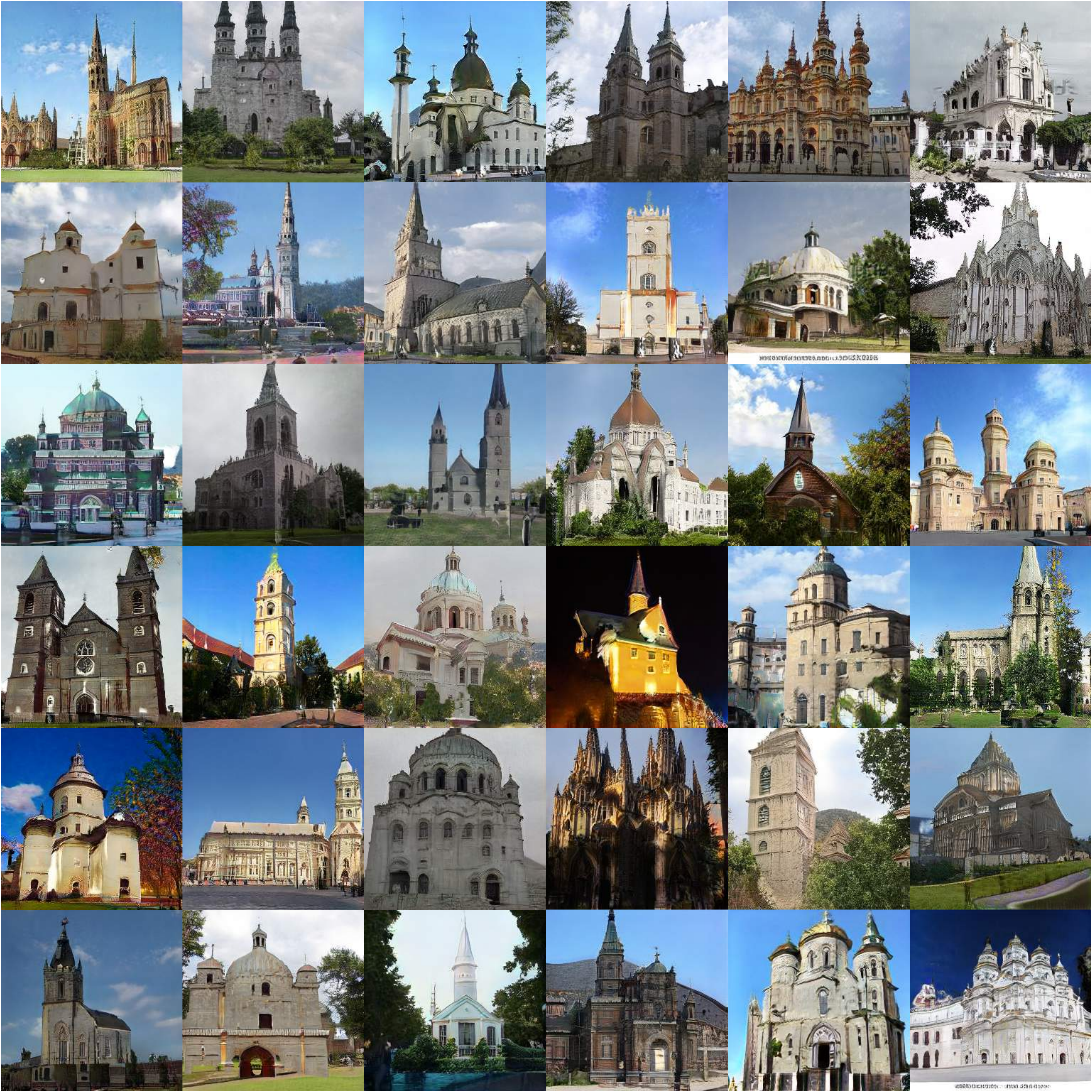}
    \caption{\textbf{Our samples.} GIRAFFE HD samples on LSUN Church~\cite{Yu2015LSUNCO} $256^2$.}
    \label{fig:sample4}
\end{figure*}

\begin{figure*}[t!]
    \centering
    \includegraphics[width=.99\textwidth]{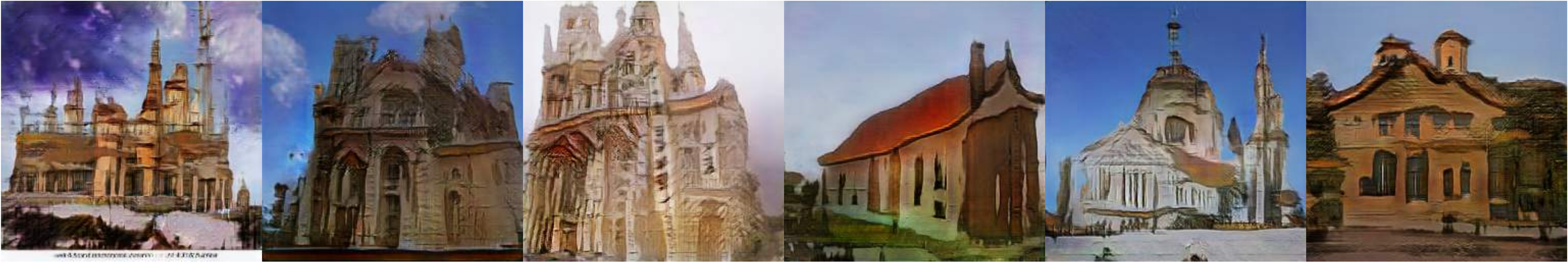}
    \caption{\textbf{GIRAFFE~\cite{Niemeyer2020GIRAFFE} samples.} GIRAFFE samples on LSUN Church $256^2$.}
    \label{fig:grf_sample3}
\end{figure*}

% \label{sec:intro}
In Figs.~\ref{fig:change_bg1} to~\ref{fig:grf_sample3}, we show additional qualitative results on controllable scene generation on four datasets: CompCar~\cite{compcars}, FFHQ~\cite{karras-cvpr2019}, AFHQ Cat~\cite{Starganv2}, LSUN Church~\cite{Yu2015LSUNCO}. Since the results on CelebA-HQ \cite{Karras-iclr2018} are very similar to those on FFHQ, we do not show the CelebA-HQ results here. We also include GIRAFFE samples on the four datasets to enable direct comparison with our method. We show the highest resolution models that we've trained for each dataset: CompCar at $512^2$, FFHQ at $1024^2$, AFHQ Cat at $256^2$, and LSUN Church at $256^2$.
% to conserve space

\begin{table*}[h]
  \centering
  \resizebox{0.99\textwidth}{!}{
  \begin{tabular}{@{}lccccccccc@{}}
    \toprule
                                      & Number of Images & Object Rotation Range     & Background Rotation Range         & Camera Elevation Range      & Horizontal Translation & Depth Translation & Vertical Translation & Object Scale & Field of View\\
    \midrule
    CompCar~\cite{compcars}            & 136,726 & $360^\circ$ & $0^\circ$  & $10^\circ$ & -0.12 - 0.12 & -0.22 - 0.22 & -0.06 - 0.08 & 0.8 - 1 & $10^\circ$\\
    FFHQ~\cite{karras-cvpr2019}        & 70,000  & $70^\circ$  & $0^\circ$  & $10^\circ$ & - & - & - & - & $10^\circ$\\
    AFHQ Cat~\cite{Starganv2}          & 5,558   & $70^\circ$  & $0^\circ$  & $10^\circ$ & - & - & - & - & $10^\circ$\\ 
    LSUN Church~\cite{Yu2015LSUNCO}    & 126,227 & $360^\circ$ & $0^\circ$  & $0^\circ$  & -0.15 - 0.15 & -0.15 - 0.15 & - & 0.8 - 1 & $30^\circ$\\
    CelebA-HQ~\cite{Karras-iclr2018}   & 30,000  & $90^\circ$  & $90^\circ$ & $10^\circ$ & - & - & - & - & $10^\circ$\\
    \bottomrule

  \end{tabular}
  }
  \caption{\textbf{Dataset parameters.} We report relevant parameters for all datasets.}
  \label{tab:params}
  \vspace{-0.1in}
\end{table*}

\end{document}